\documentclass[letterpaper, 10 pt, conference]{ieeeconf}  % Comment this line out if you need a4paper

\IEEEoverridecommandlockouts                              % This command is only needed if 
\usepackage{graphics} % for pdf, bitmapped graphics files
\usepackage{epsfig} % for postscript graphics files
\usepackage{mathptmx} % assumes new font selection scheme installed
\usepackage{times} % assumes new font selection scheme installed
\usepackage{amsmath} % assumes amsmath package installed
\usepackage{amssymb}  % assumes amsmath package installed
\usepackage{xspace}
\usepackage[caption=false]{subfig}
\usepackage{threeparttable}
\usepackage{url}
\usepackage{multirow}
\usepackage{booktabs} 
\usepackage{soul}
\title{\LARGE \bf
AIR-HLoc: Adaptive Retrieved Images Selection\\ for Efficient Visual Localisation
}

\author{Changkun Liu$^{1}$, Jianhao Jiao$^{2}$, Huajian Huang$^{1}$, Zhengyang Ma$^{3}$, Dimitrios Kanoulas$^{2,5}$ and Tristan Braud$^{1,4}$% <-this % stops a space
\thanks{$^{1}$Department of Computer Science and Engineering, HKUST, Hong Kong, {\tt\small \{cliudg, hhuangbg\}@connect.ust.hk}. Corresponding author: Tristan Braud, {\tt\small braudt@ust.hk}}%
\thanks{$^{2}$ Department of Computer Science, University College London, London, The United Kingdom, {\tt\small \{ucacjji, d.kanoulas\}@ucl.ac.uk}}
\thanks{$^{3}$Division of Emerging Interdisciplinary Areas, HKUST, Hong Kong, {\tt\small zmaaf@connect.ust.hk}
}%
\thanks{$^{4}$Division of Integrated Systems Design, HKUST, Hong Kong.}%
\thanks{\textsuperscript{5}Dimitrios Kanoulas is also with the AI Centre, Department of Computer Science, University College London, Gower Street, WC1E 6BT, London, UK and Archimedes/Athena RC, Greece.}
\thanks{This work were supported, in part, by HKUST OKT Project No. FTRIS-23-015, 
ITF Project No. ITS/319/22FP, and  
the UKRI Future Leaders Fellowship [MR/V025333/1] (RoboHike).}
}%

\newcommand{\sysname}{AIR-HLoc\xspace}
\begin{document}

\maketitle
\thispagestyle{empty}
\pagestyle{empty}

%%%%%%%%%%%%%%%%%%%%%%%%%%%%%%%%%%%%%%%%%%%%%%%%%%%%%%%%%%%%%%%%%%%%%%%%%%%%%%%%
\begin{abstract}
State-of-the-art hierarchical localisation pipelines (HLoc) employ image retrieval (IR) to establish 2D-3D correspondences by selecting the top-$k$ most similar images from a reference database. While increasing $k$ improves localisation robustness, it also linearly increases computational cost and runtime, creating a significant bottleneck. This paper investigates the relationship between global and local descriptors, showing that greater similarity between the global descriptors of query and database images increases the proportion of feature matches. Low similarity queries significantly benefit from increasing $k$, while high similarity queries rapidly experience diminishing returns. Building on these observations, we propose an adaptive strategy that adjusts $k$ based on the similarity between the query's global descriptor and those in the database, effectively mitigating the feature-matching bottleneck. Our approach reduces computational costs and processing time without sacrificing accuracy. Experiments on three indoor and outdoor datasets show that \sysname reduces feature matching time by up to  30\% while preserving state-of-the-art accuracy. The results demonstrate that \sysname facilitates a latency-sensitive localisation system.

\end{abstract}

\section{INTRODUCTION}
Visual localisation systems estimate 6 degrees of freedom (6DoF) absolute camera poses for query images. Accurate and real-time absolute 6DoF pose estimation is crucial in diverse applications, from augmented reality to navigation and decision-making for mobile robots. Standard structure-based methods~\cite{sattler2012image,dusmanu2019d2,sarlin2019coarse,taira2018inloc,germain2019sparse,sarlin2022lamar, sarlin2020superglue} achieve high accuracy by detecting and matching visual features in query images with a 3D representation of the environment. To reduce the search space in large datasets, standard hierarchical localisation pipelines (HLoc)~\cite{sarlin2019coarse, sarlin2020superglue} include an image retrieval (IR) module to establish 2D–2D matches between the query image and the top-$k$ ranked database images. The $k$ retrieved images are used to build 2D-3D correspondences based on a smaller subset of the environment's 3D model. Camera pose is finally estimated via RANSAC and Perspective-n-Point (PnP) algorithms~\cite{kneip2011novel,fischler1981random}. 

\begin{figure}[t]
  \centering
  \includegraphics[width=0.9\linewidth]{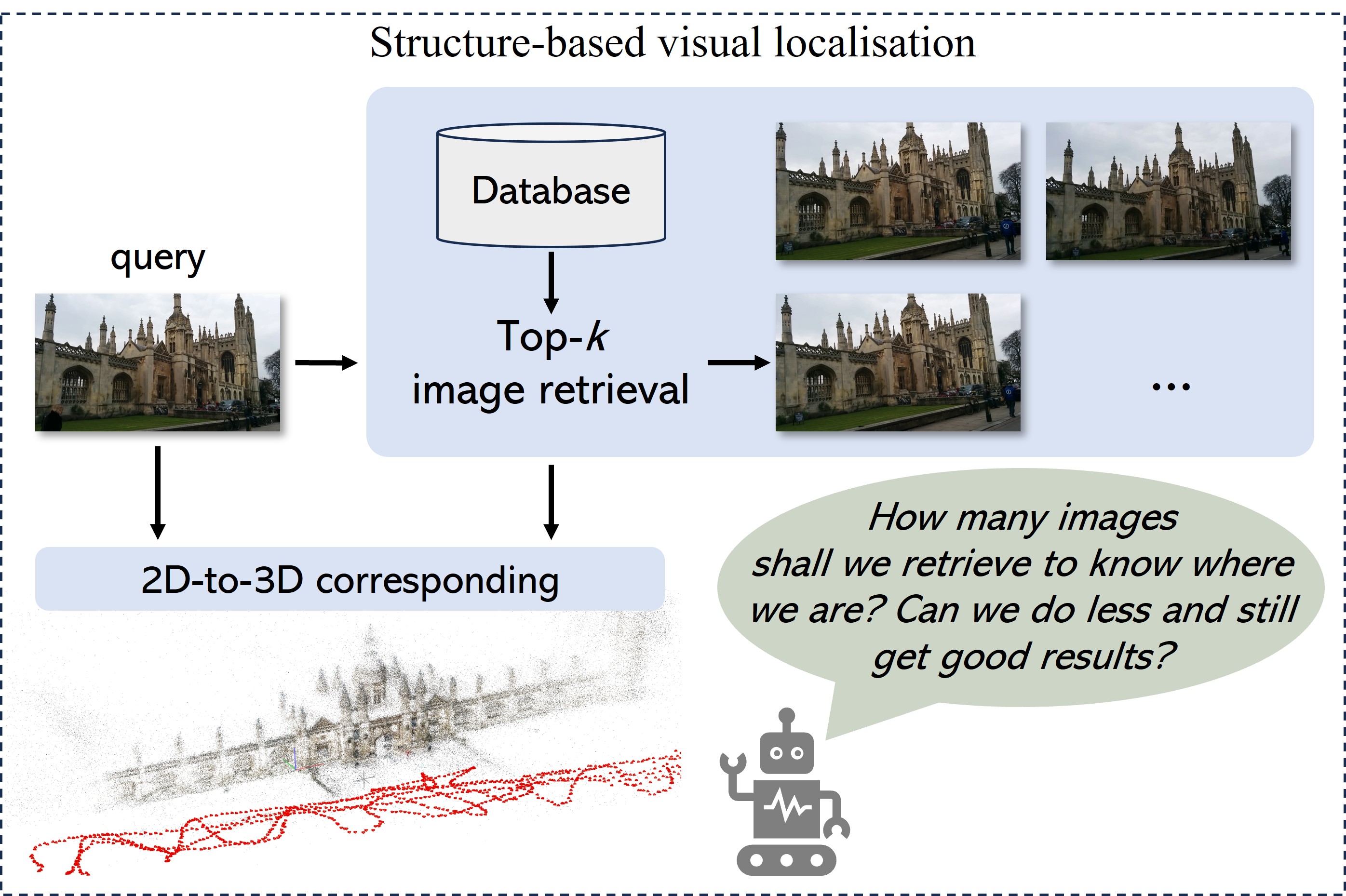}
  \caption{AIR-HLoc proposes an efficient yet practical solution to enhance localisation efficiency while maintaining accuracy by adaptively retrieving a varying number of images for different queries.}
\label{fig:hloc}
\end{figure}

These methods typically use a fixed value of $k$, determined empirically to balance robustness and performance for a given dataset~\cite{humenberger2022investigating}. The choice of $k$ is crucial, as it directly impacts localisation accuracy and runtime. This work seeks to optimize $k$ selection to minimize runtime without compromising localisation accuracy A larger $k$ increases the number of potential matches from similar images, enhancing localisation robustness for challenging queries and improving overall positioning accuracy. However, it also extends feature matching time, becoming a key bottleneck in the pipeline, and may introduce noisy matches. Conversely, a smaller $k$ reduces runtime and improves efficiency, but at the cost of localisation accuracy. Previous works~\cite{humenberger2022investigating, sarlin2022lamar} recommend using a moderate value of $k$, typically between 5 and 30. However, applying a fixed $k$ for all queries can reduce accuracy (if $k$ is too small) or waste computational resources and processing time (if $k$ is too large).

To address this challenge, we explore strategies for achieving a better trade-off by selecting an appropriate $k$ for each query in real-world latency-sensitive visual localisation systems, as shown in Figure~\ref{fig:hloc}. Since global descriptors used in IR methods provide a high-level image representation, we calculate the cosine similarity between the global descriptors of the query image and the top-$3$ retrievals. This serves as the basis for the scoring mechanism described in Section~\ref{sec:method}, which quantifies the query image's similarity to the reference database images. Our observations indicate that query images with varying scores exhibit different levels of localisation difficulty. Higher scores, representing greater similarity and more local feature matches, correspond to easier localisation, where fewer images are required to establish sufficient 2D-3D correspondences.
 we introduce an \textit{\underline{a}dapt\underline{i}ve \underline{r}etrieved} image selectionp two benefits. Firsthe, it allows us to resrieve fewer images while, which adaptivelocadjust accurnd, thore of each query reflects the uncertainty for different queries of posmethod reduces the number ofr of retrdimages  and feature matching costsfohile maintain of differsnt difficultie.
%, and it provides a measure ofilisation accurayuncertainty for each query.

%Our experiments demonstrate that, even with a sufficiently large $k$, the average pose estimation error for images with low scores significantly exceeds that for images with high scores.

We summarize our main contributions as follows:
\begin{enumerate}
\item We begin with a statistical analysis of the relationship between global and local descriptors. Across three indoor and outdoor datasets, using three mainstream IR models, we observe that higher image similarity correlates with a greater proportion of feature matches.

\item We propose using the cosine similarity of global descriptors to assess the localisation difficulty of queries. Building on this, we introduce an \textit{\underline{a}dapt\underline{i}ve \underline{r}etrieved} image selection (AIR) approach for HLoc. For easier queries, we retrieve fewer images, while for more challenging queries, we retrieve more images. Additionally, we present a novel metric, \textit{Mean
Localisation Improvement Per-retrieved-image (MLIP)}, $\zeta$, to quantify the improvement in localisation accuracy as $k$ varies across different queries. This insight facilitates more informed $k$ selection and algorithm design in practical applications.
\item Our \sysname reduces the average matching time by 30\%, 26\%, and 11\% across Cambridge Landmarks, 7Scenes, and Aachen Day-Night-v1.1 datasets, respectively, while maintaining state-of-the-art accuracy. %The experimental settings and sample size further validate the generalizability of our approach. 
On the Cambridge dataset, \sysname reduces the average query runtime by 2433 ms at $k = 10$ on the Jetson Orin platform, while achieving higher accuracy than HLoc.

%Specifically, on the Cambridge dataset, the average runtime of all queries is reduced by 360 ms to 720 ms when $k=10\sim20$. For easier queries, \sysname reduces runtime by 600 ms to 1200 ms when $k=10\sim20$. The experimental settings and sample size further validate the generalizability of our findings and approach.
\end{enumerate}

\section{Related Work}
\subsection{Visual localisation}
\textbf{Structure-based methods} estimate camera poses by establishing matches between the 2D features of the query image and 3D points of pre-built 3D models. The matches can be established indirectly with local feature extraction and matching~\cite{sarlin2019coarse,lowe2004distinctive,sarlin2020superglue,detone2018superpoint,tyszkiewicz2020disk} or directly with scene coordinate regression~\cite{brachmann2017dsac,brachmann2021dsacstar,brachmann2018lessmore,brachmann2023accelerated,wang2024glace}. However, even though scene coordinate regression (SCR) methods achieve high accuracy in small and medium indoor scenes, they are not robust and accurate enough for larger and more complex scenes~\cite{brachmann2019expert,brachmann2018lessmore,li2020hierarchical}. 
HLoc pipeline~\cite{sarlin2019coarse,sarlin2020superglue,he2024accurate} achieves state-of-the-art (SOTA) accuracy on both indoor and large-scale outdoor scenes by using IR~\cite{arandjelovic2016netvlad, ge2020self, GARL17} as an intermediate step, allowing it to scale up to larger scenes.

\textbf{Absolute Pose Regressors} (APRs) are end-to-end learning-based methods that directly regress the absolute camera pose from input images. PoseNet~\cite{kendall2015posenet,kendall2017geometric} and the follower APRs~\cite{shavit2021learning,chen2021direct,chen2022dfnet,brahmbhatt2018geometry,moreau2022coordinet,chen2024map} 
can provide faster predictions than structure-based methods. However, they have low accuracy and robustness and are very difficult to generalize well to novel viewpoints, which are very different from the training set~\cite{sattler2019understanding,liu2024hr}. 
Therefore, this paper focuses on the HLoc pipeline, the indirect structure-based approach with IR, because it can provide robust and accurate results in large-scale scenes. 

\subsection{Image Retrieval}
Image Retrieval (IR) is commonly used for place recognition and visual localisation tasks. IR fetches top-$k$ relevant and similar database images given query images by extracting global descriptors from pre-trained models~\cite{arandjelovic2016netvlad, GARL17, Berton_2023_EigenPlaces,yin2024general}. We focus on IR in structure-based localisation in this paper.
~\cite{humenberger2022investigating} studies the performance correlation between classical place recognition and localisation tasks and finds almost no improvement in visual localisation when $k$ is larger than 20. Mobile sensors' data are utilized to reduce the database's research space when doing IR in~\cite{yan2023long}.~\cite{humenberger2007robust,sarlin2022lamar} use a fusion of global descriptors in the visual localisation task.
This study aims to reduce computational costs and processing time of HLoc by adaptively allocating the number of retrieved images for more efficient matching instead of using a constant $k$ for all queries in previous work.
\section{Methods}
\label{sec:motivation}
Given a query image $I^q \in \mathbb{R}^{H \times W \times 3}$, HLoc predicts a 6DoF camera pose $\hat{p}_k =  [\mathbf{\hat{x}}_k,\mathbf{\hat{q}}_k]$ by retrieving the top-$k$ similar images to $I^q$. $\mathbf{\hat{x}}_k\in \mathbb{R}^3$ denotes the estimated global translation, $\mathbf{\hat{q}}_k \in \mathbb{R}^4$ denotes the estimated quaternion, which encodes the rotation. In this section, we analyze the impact of the hyperparameter \( k \) on the runtime and localisation accuracy for different query images. Based on our findings, we propose a simple yet effective scheme, \sysname, to enhance the efficiency of HLoc while maintaining the accuracy.

\subsection{Problem Formulation}
In the HLoc pipeline, given a query image \( I^q \), an image retrieval (IR) model retrieves the top-\( k \) images from a reference database \( D \), ranked by the cosine similarity of their global descriptors, as illustrated in Figure~\ref{fig:hloc}. HLoc then establishes 2D-2D correspondences between the query image \( I^q \) and the retrieved images utilizing feature matching. By combining these 2D-2D correspondences with the 3D scene model, HLoc establishes 2D-3D correspondences between \( I^q \) and the scene, and predicts a 6DoF camera pose \( \hat{p}_k = [\mathbf{\hat{x}}_k, \mathbf{\hat{q}}_k] \).

Feature matching is the bottleneck of Hloc. The larger the $k$, the longer it takes to perform feature matching. We show the detailed data in Section~\ref{subsec:se}. We aim to reduce the feature-matching cost by retrieving top-$k^*$ reference images, $k^* \leq k$ for the different query image $I^q$ while maintaining robustness and accuracy. We first investigate several mainstream image retrieval methods, including AP-GeM~\cite{GARL17}, NetVLAD~\cite{arandjelovic2016netvlad}, and EigenPlaces~\cite{Berton_2023_EigenPlaces}. We aim to show that feature matching strongly correlates with the global visual similarity between query and reference database images. Furthermore, different queries benefit to varying degrees from increasing $k$ in terms of accuracy. So we argue that existing methods of retrieving an equal number of images for all queries are a waste of time and computational power.

\subsection{Relationship between \textit{match ratio} and similarity}
Given a query image $I^q$, the image database is $D$.  An IR model $E$ extracts the global descriptors of the query image $I^q$ as $g^q$. $E$ extracts the global descriptors of reference images $I^r, 
 I^r \in D$ as $g^r$. The similarity between $I^q$ and $I^r$ is 
 \begin{equation}
    \cos(g^{q},g^{r} ) = \frac{g^{q} \cdot g^{r} }{||g^{q}||_2\cdot||g^{r}||_2}.
\label{eq:cosimi}
\end{equation}
 Similarly, a local feature extraction model $F$ extracts the local feature of the query image $I^q$ as $f^q$. $F$ extracts the local feature of $I^r$ as $f^r$. The number of local feature points of $I^q$ is $N(f^q)$, and the number of matched local features between $f^r$ and $f^q$ is $M(f^q, f^r)$. We define the \textit{match ratio} as:
 \begin{equation}
     \text{\textit{match ratio}} = \frac{M(f^q, f^r)}{N(f^q)}, 
\label{eq:match_ratio}
 \end{equation}
where $0\leq \text{\textit{match ratio}}\leq 1$. In this paper, we use the prevalent SuperPoint (SP)~\cite{detone2018superpoint} and SuperGlue (SG)~\cite{sarlin2020superglue} as the feature extractor and matcher.

\begin{figure}
\centering
  \subfloat[Aachen, AP-GeM\\ (PCC:0.63, SRC:0.62)]{
  %% label for second subfigure
 \includegraphics[width=.32\linewidth]{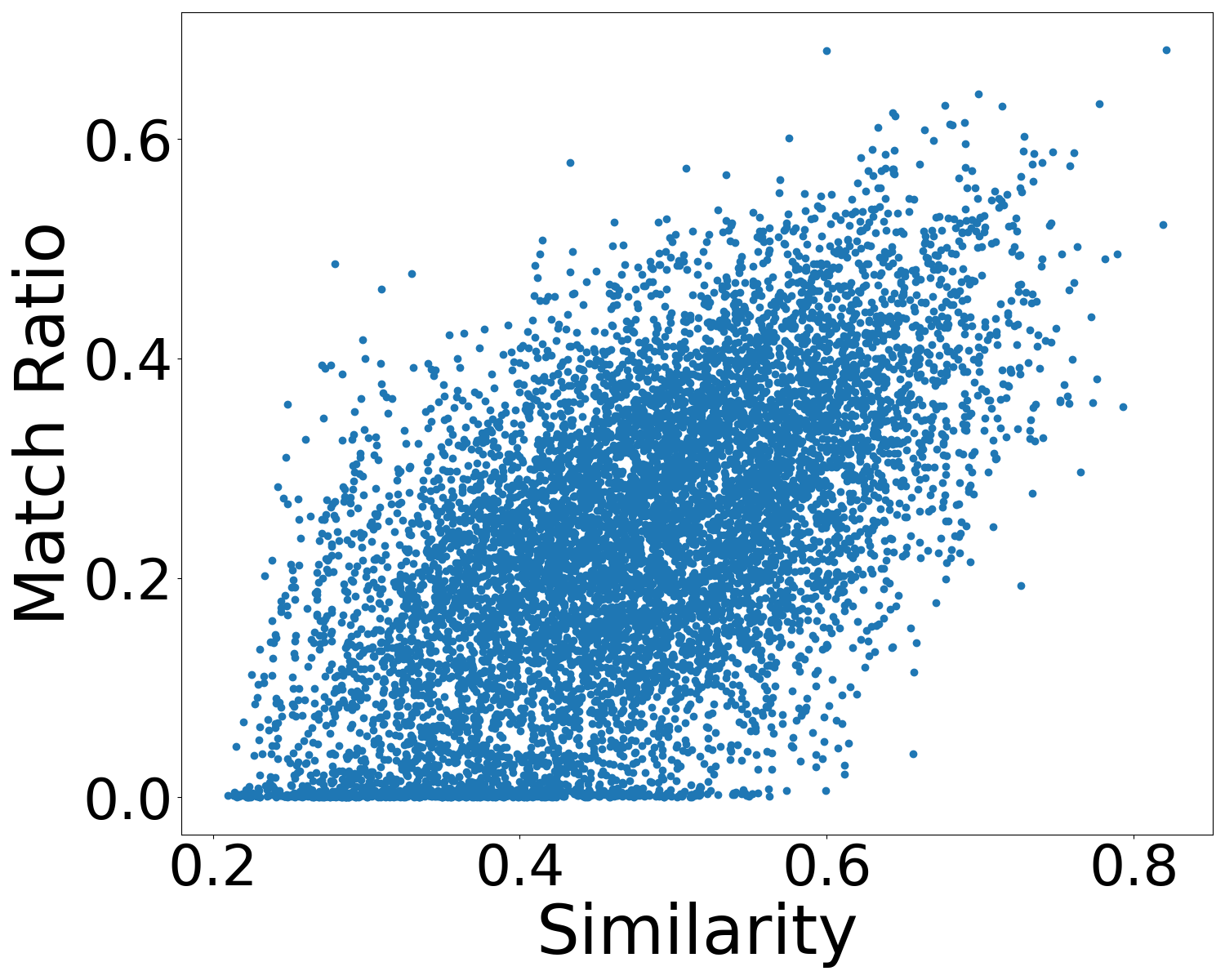}}
  \subfloat[Aachen, NetVLAD\\(PCC:0.68, SRC:0.67) ]{
  %% label for second subfigure
 \includegraphics[width=.32\linewidth]{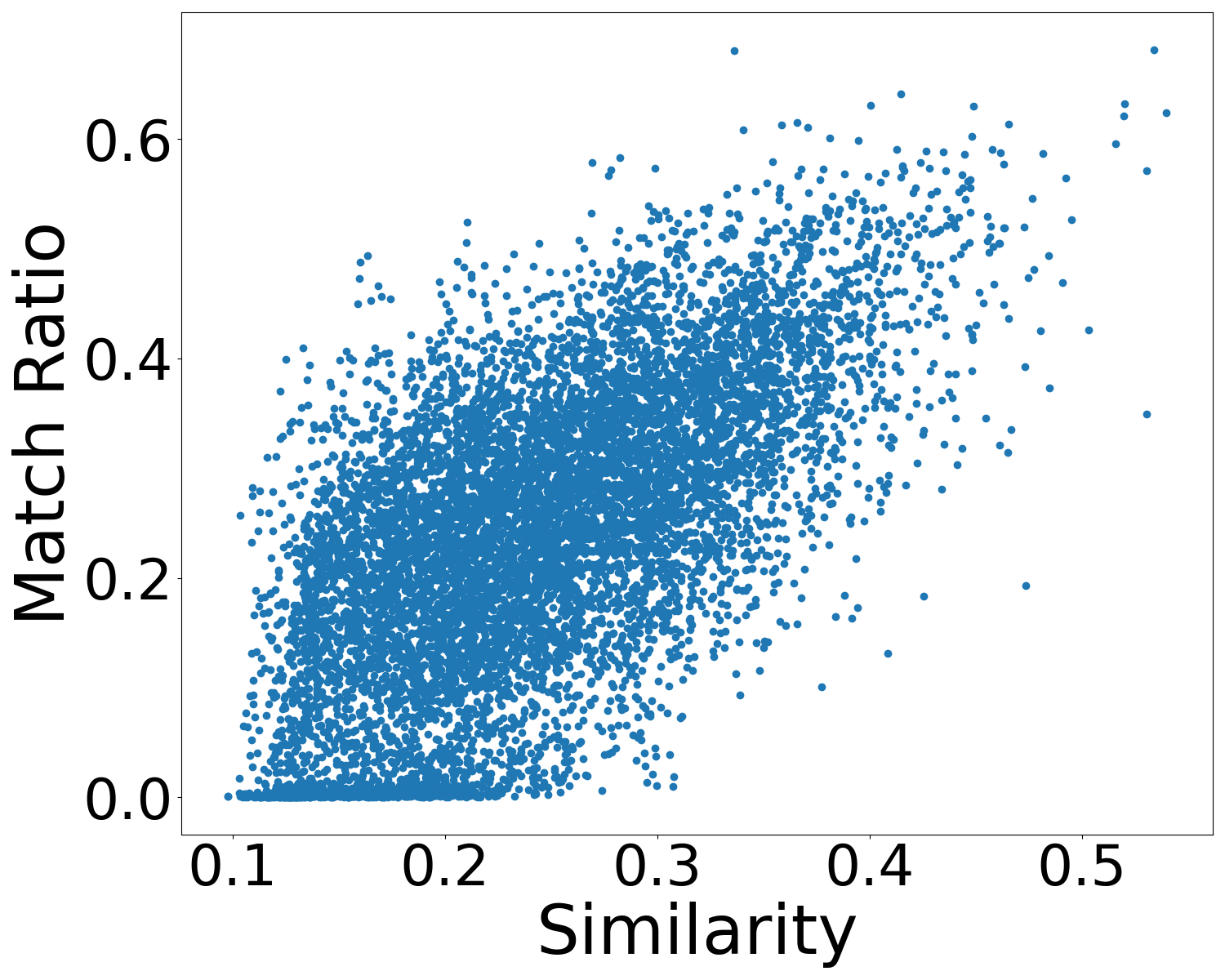}}
 \subfloat[Aachen, EigenPlaces\\(PCC:0.72, SRC:0.71)]{
  %% label for second subfigure
 \includegraphics[width=.32\linewidth]{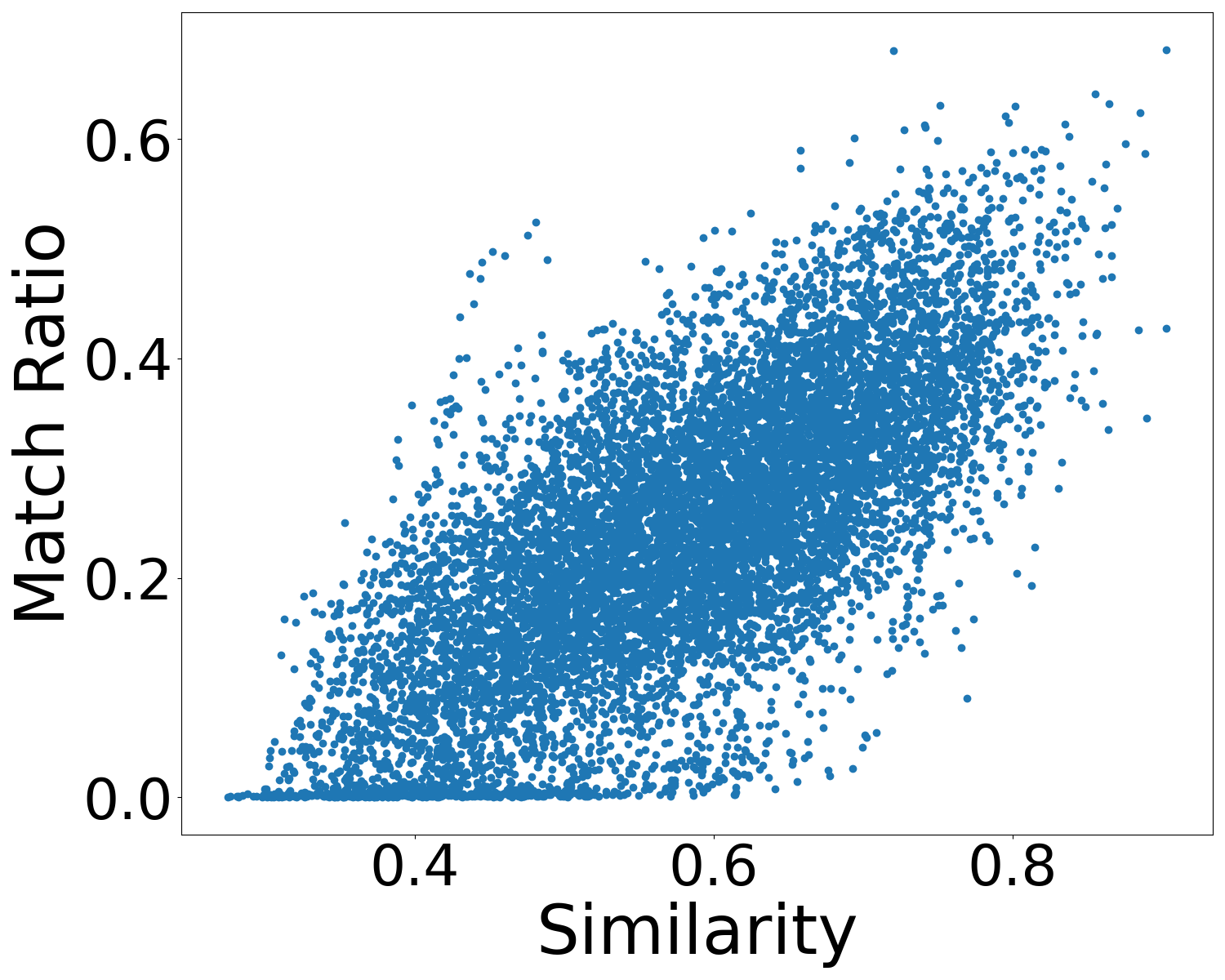}}
 \newline
 \includegraphics[width=.98\linewidth]{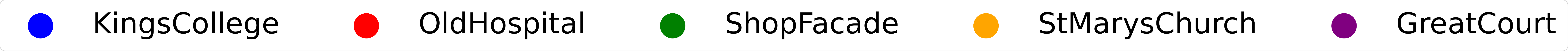}
\vspace{-1em} % 

\subfloat[Cam., AP-GeM\\ (PCC:0.61, SRC:0.58)]{
 %% label for second subfigure
\includegraphics[width=.32\linewidth]{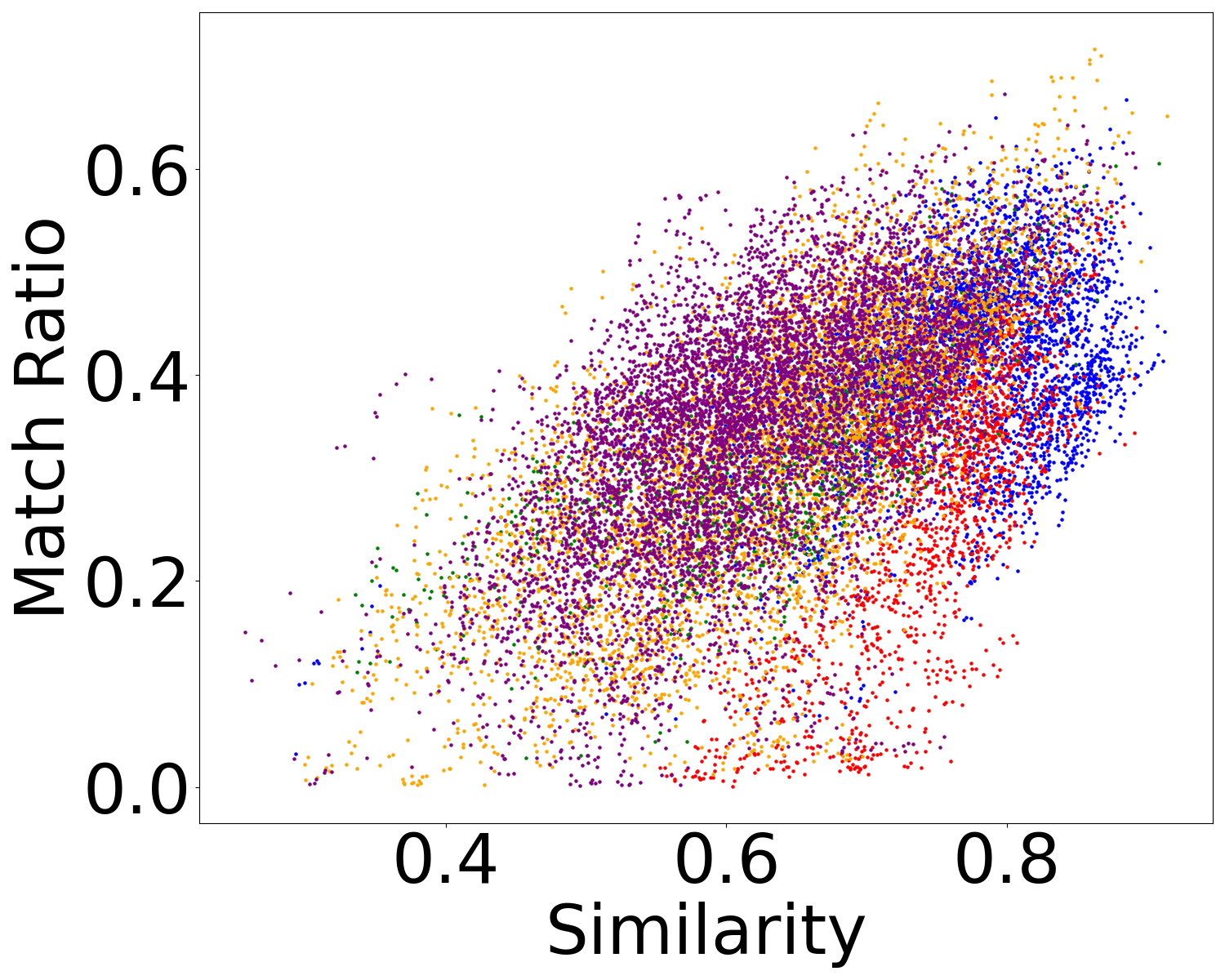}}
\subfloat[Cam., NetVLAD\\ (PCC:0.67, SRC:0.65)]{
 %% label for second subfigure
\includegraphics[width=.32\linewidth]{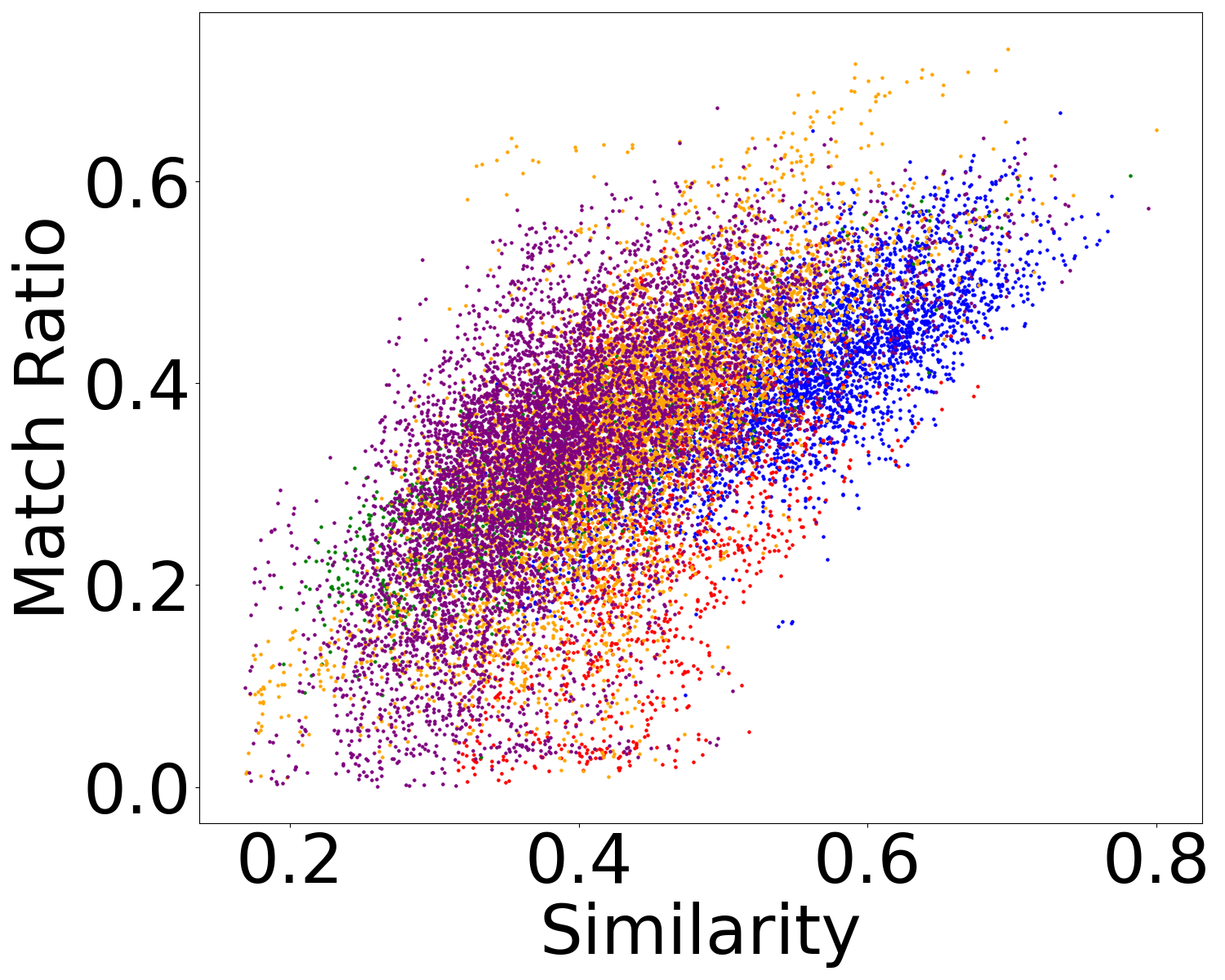}}
\subfloat[Cam., EigenPlaces\\(PCC:0.67, SRC:0.65)]{
 %% label for second subfigure
\includegraphics[width=.32\linewidth]{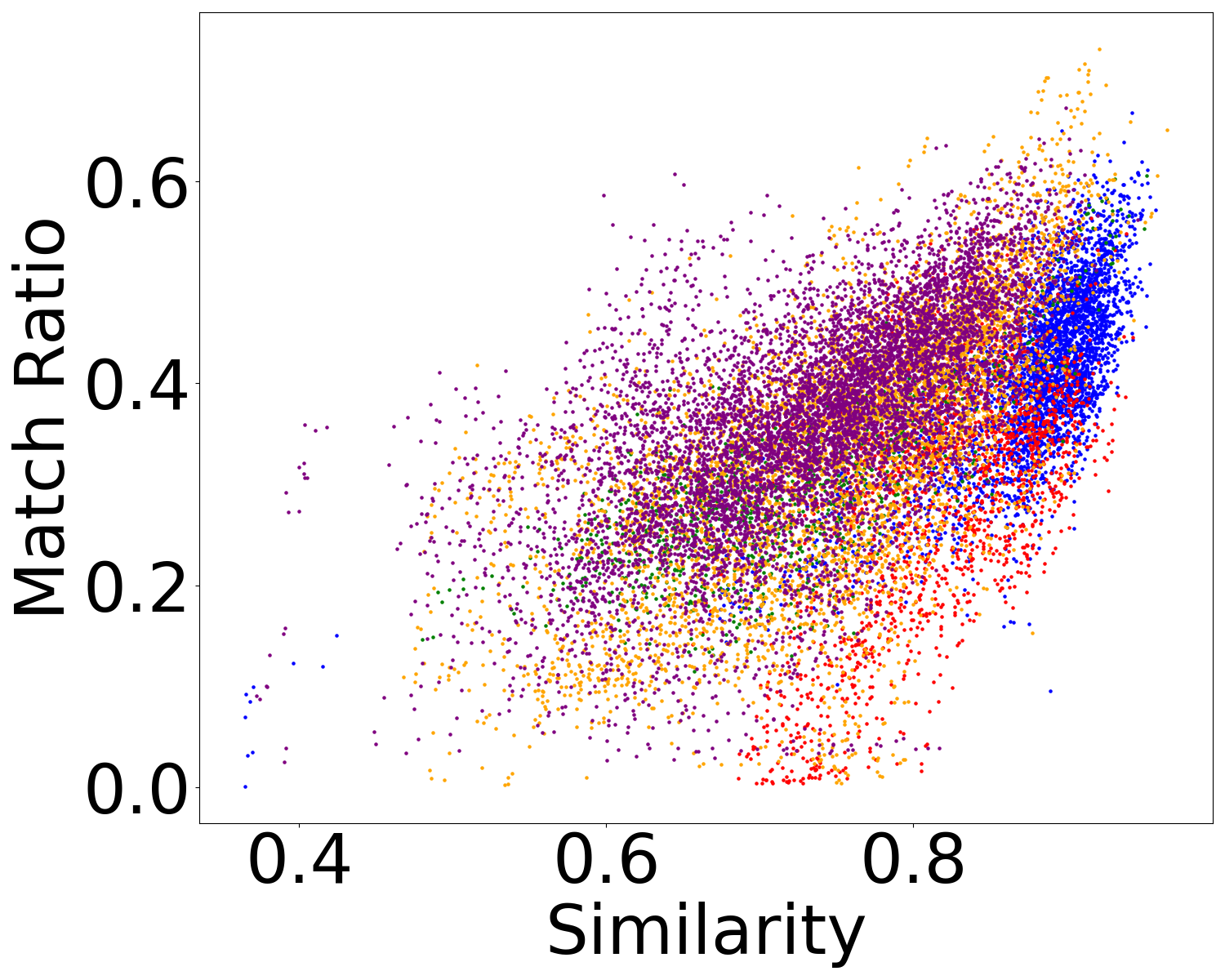}}
\newline
\includegraphics[width=.98\linewidth]{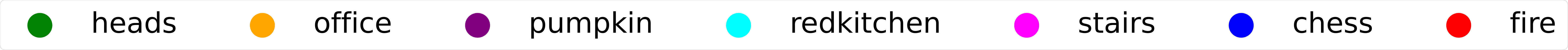} 
\vspace{-0.9em} % 

\subfloat[7S., AP-GeM\\ (PCC:0.51, SRC:0.48)]{
  %% label for second subfigure
 \includegraphics[width=.32\linewidth]{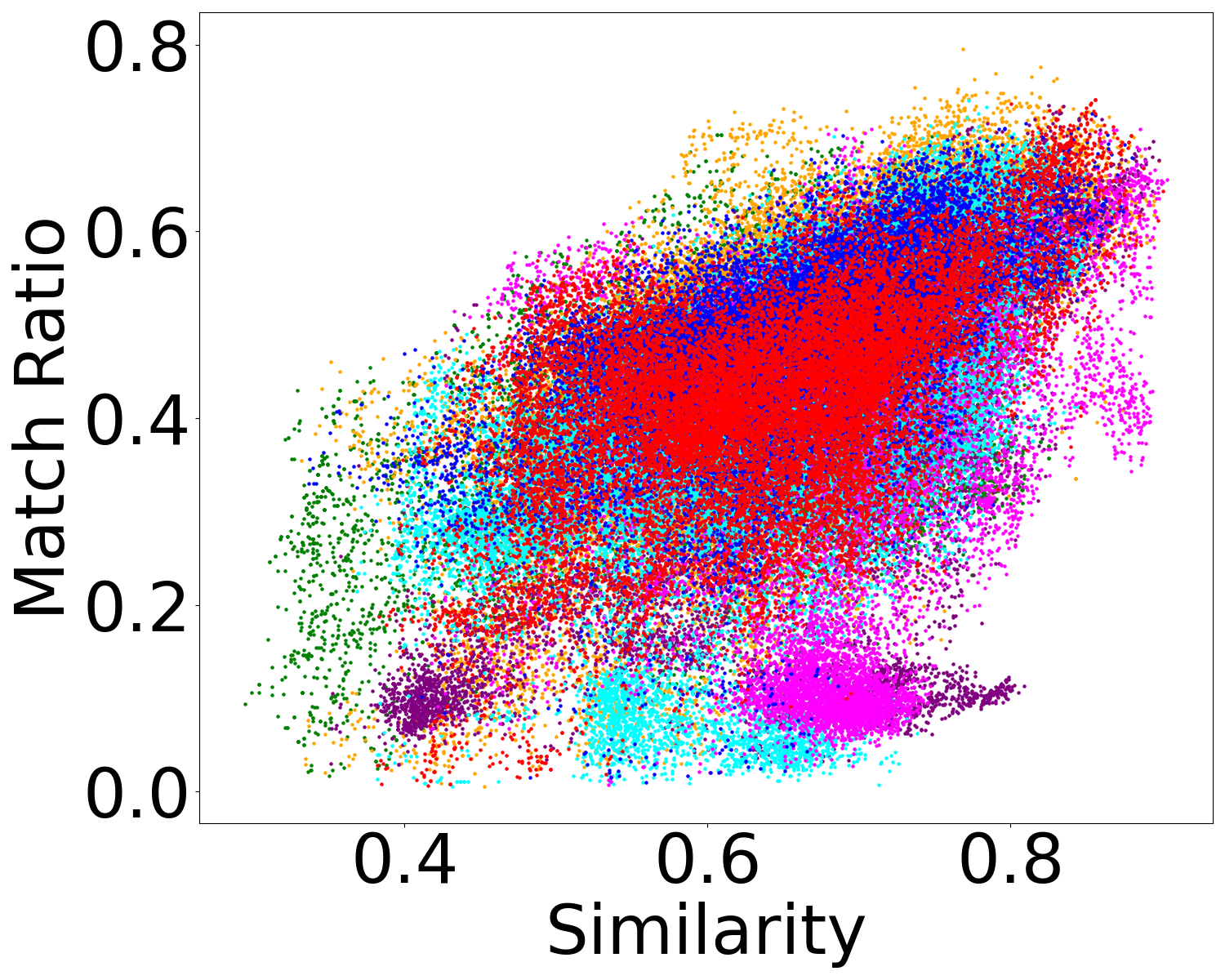}}
\subfloat[7S., NetVLAD\\ (PCC: 0.61, SRC:0.58)]{
  %% label for second subfigure
 \includegraphics[width=.32\linewidth]{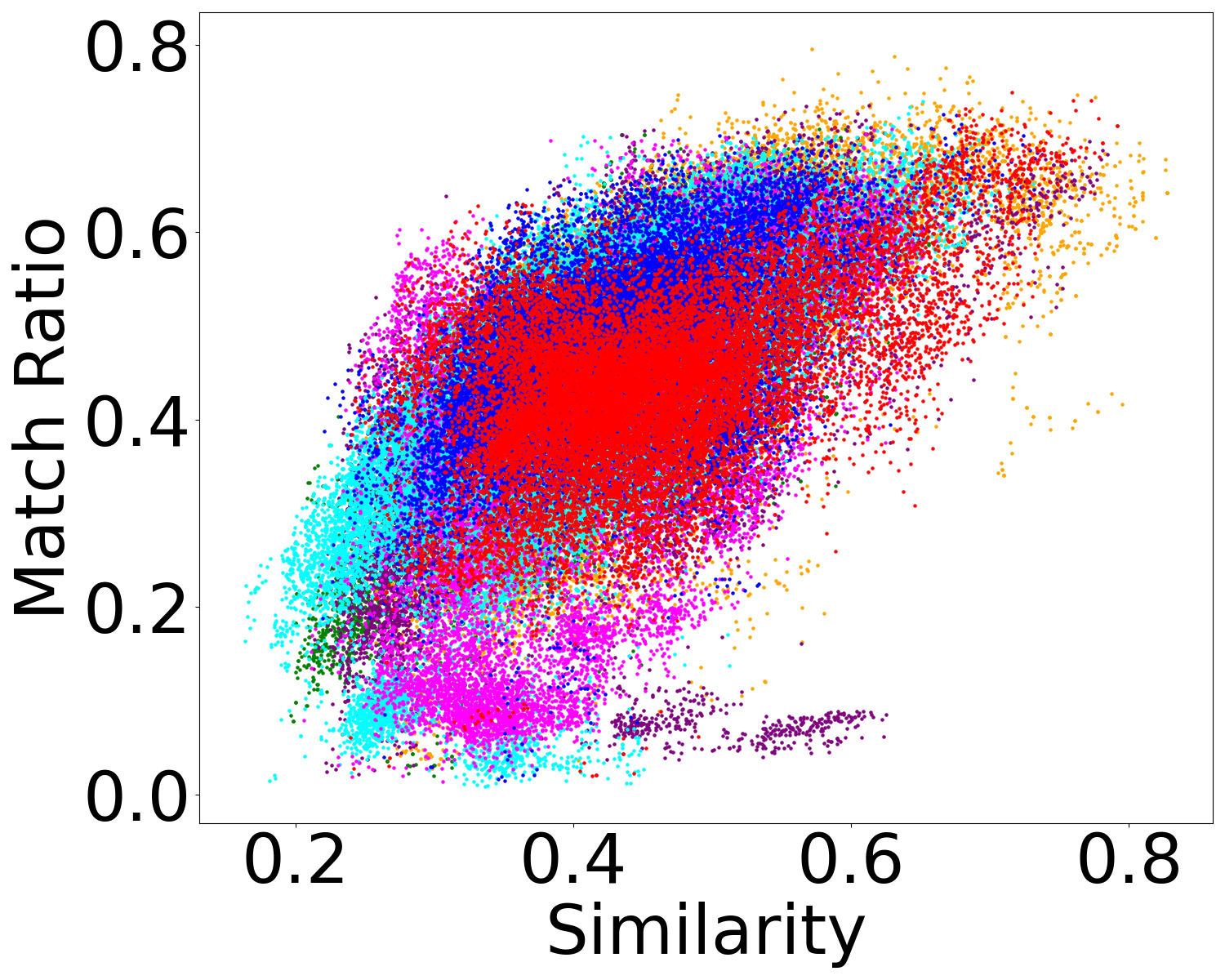}}
\subfloat[7S., EigenPlaces\\ (PCC:0.51, SRC:0.51)]{
  %% label for second subfigure
 \includegraphics[width=.32\linewidth]{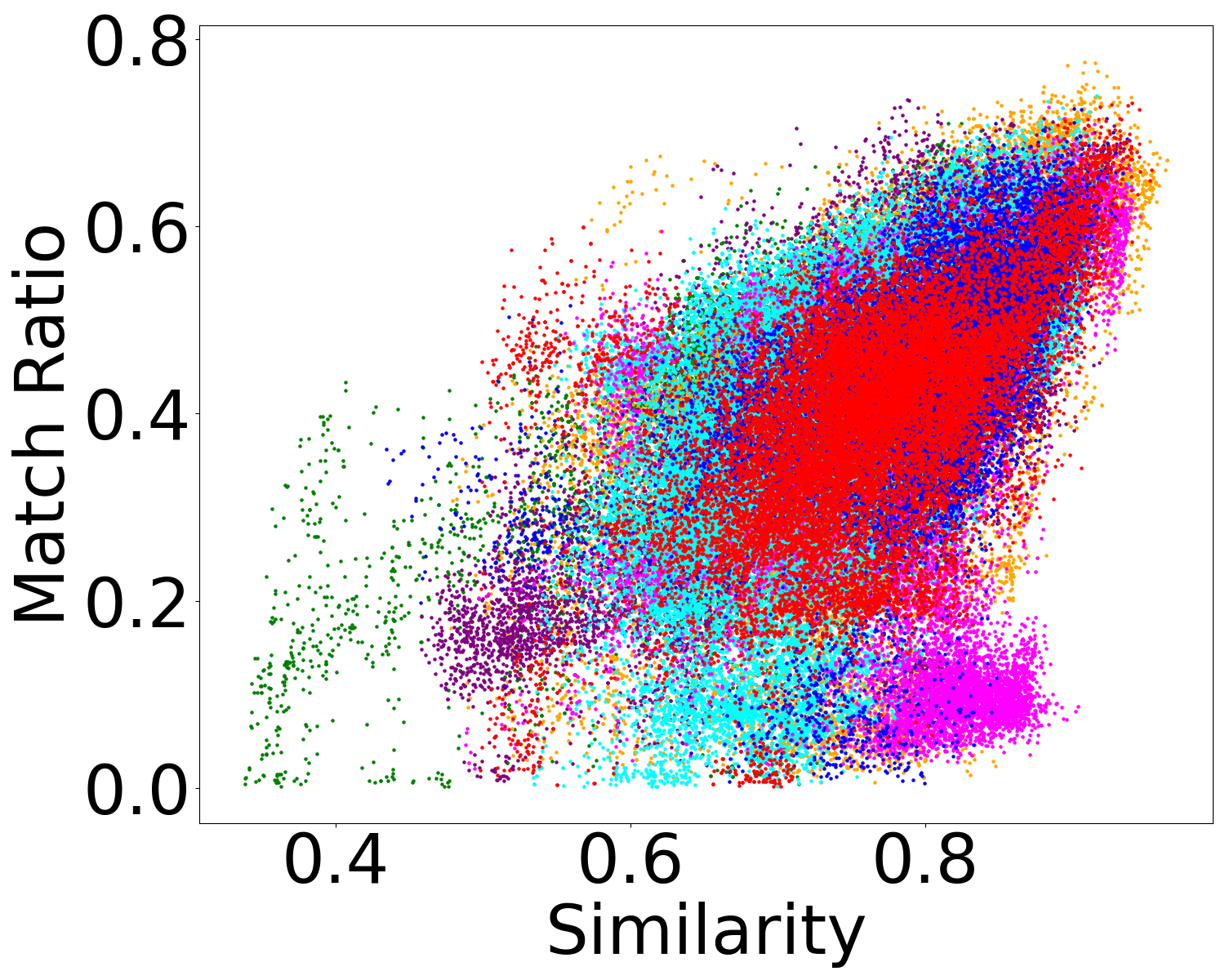}}
 \newline
 \caption{Subfigures (a)-(f) show the correlation between cosine similarity (-1 to 1) and \textit{match ratio} (0 to 1) in Aachen Day-Night-v1.1 datasets~\cite{sattler2018benchmarking,zhang2021reference}, Cambridge landmark~\cite{kendall2015posenet},  and 7Scenes~\cite{glocker2013real,shotton2013scene} using three IR models (AP-GeM, NetVLAD and EigenPlaces). The value in parentheses of each subfigure is the average PCC and SRC across all scenes in a dataset.} %PCC values closer to 1 indicate a stronger linear positive correlation, while SRC values closer to 1 indicate a stronger positive monotonic correlation.
 \label{fig:match_ratio_simi} %% label for entire figure
\end{figure}

Figure~\ref{fig:match_ratio_simi} shows the strong correlation between cosine similarity in Equation~\ref{eq:cosimi} and \textit{match ratio}, as described in Equation~\ref{eq:match_ratio}, for each $I^q$ and $I^r$ pair. The number in parentheses of each subplot is the Pearson correlation coefficient (PCC)~\cite{pearson1895vii} and Spearman correlation coefficient (SRC)~\cite{myers2013research}, both ranging from -1 to 1. A larger PCC value indicates a stronger linear positive correlation, while an SRC value close to 1 suggests a strong positive monotonic relationship. We opt to use the \textit{match ratio} instead of the absolute number of matches in our analysis, as some query images exhibit textureless regions with fewer feature points, yielding only a few matches even when highly similar images are retrieved. Utilizing a \textit{match ratio} from 0 to 1 provides a more meaningful measure. Our analysis incorporates three distinct IR models (AP-GeM, NetVLAD, and EigenPlaces). AP-GeM and NetVLAD serve as representative IR models in the visual localisation pipeline~\cite{sarlin2022lamar,sarlin2019coarse,humenberger2022investigating}. A strong positive correlation between \textit{match ratio} and similarity is observed for all three IR models across three different datasets. Therefore, if a query image can find very similar images, then a small number of retrieved images can contribute enough matches for pose estimation. For queries that have difficulty finding similar images, we need more retrieved images.

\subsection{Adaptive Retrieved Images Selection}
\label{sec:method}
According to the correlation observed in Figure~\ref{fig:match_ratio_simi}, we propose to use the similarity between the images in the database and the query to evaluate the localisation difficulty of query images. We first store the global descriptors $G^r$ of all the images in the reference images database offline. For a query image $I^q$, we get the global descriptor $g^q$ using a pre-trained IR model. Then, we calculate the cosine similarity between $g^q$ and each $g^r \in G^r$.
The similarity is then ranked from highest to lowest, with the mean of the top-$n$ being the score $S(I^q)$ of the $I^q$. 
\begin{equation} S(I^q) = \frac{1}{n} \sum\limits_{j \in J} \cos(g^{q}, g^{j}), 
\label{eq:score}
\end{equation}
where $J$ represents the indices of the top $n$ images in the ranked list. In this paper, we set $n = 3$. We believe that the three most similar images are sufficient to reflect the localisation difficulty of this query. Therefore, instead of retrieving fixed $k$ images for all queries, we retrieve different numbers of reference images for different queries depending on $S(I^q)$ in Equation (\ref{eq:score}). We denote the maximum number of images that are retrieved as $k$ for each $I^q$. Then, for different $I^q$, we follow the below rules:
\begin{itemize}
    \item For easy query, $S(I^q) \geq \gamma^h$, we retrieve top-$k^*$ images, $k^* = \lceil \alpha\times k \rceil$.
    \item For medium query, $\gamma^l \leq S(I^q) < \gamma^h$, we retrieve top-$k^*$ images, $k^* = \lceil \beta\times k \rceil$.
    \item For hard query, $S(I^q) < \gamma^l$, we retrieve top-$k$ images.
\end{itemize}
, where $\gamma^h$, $0<\gamma^h\leq1$ is the high similarity threshold.  $\gamma^l$, $0<\gamma^l < \gamma^h\leq1$ is the low similarity threshold. $0< \alpha < \beta <1$ are coefficients to reduce the number of retrieved images. $\lceil \cdot \rceil$ is the ceiling function. Since we adaptively assign the retrieved number for each $I^q$, we call our improved HLoc approach \sysname.

\subsection{Mean Localisation Improvement Per-retrieved-image}
To assess the impact of increasing \( k \) on localisation accuracy, we propose a new metric called the \textit{Mean Localisation Improvement Per-retrieved-image} (MLIP), denoted as \( \zeta = (\zeta_T(k), \zeta_R(k)) \). The term \( \zeta_T(k) \) represents the reduction in translation error relative to the baseline \( k = 1 \) for a given query image \( I^q \), while \( \zeta_R(k) \) captures the reduction in rotation error. Higher values of \( \zeta_T(k) \) or \( \zeta_R(k) \) signify a greater average contribution of the retrieved top-$k$ images to the overall improvement in localisation accuracy.

\begin{equation}
    \zeta_{T}(k) = -\frac{\text{ATE}(k) - \text{ATE}(1)}{k - 1}, \quad k \geq 2,
\end{equation}

\begin{equation}
    \zeta_{R}(k) = -\frac{\text{ARE}(k) - \text{ARE}(1)}{k - 1}, \quad k \geq 2.
\end{equation}

The absolute translation error (ATE) and absolute rotation error (ARE) are defined as follows: $\text{ATE}(k) = \| \mathbf{\hat{x}}_k - \mathbf{x} \|_2$, $\text{ARE}(k) = 2 \arccos{\left|\mathbf{q}^{-1} \mathbf{\hat{q}}_k \right|} \cdot \frac{180}{\pi}$. Here, \( \mathbf{q}^{-1} \) denotes the conjugate of the ground truth quaternion \( \mathbf{q} \), and \( p = [\mathbf{x}, \mathbf{q}] \) represents the ground truth 6DoF camera pose of \( I^q \).

\section{Evaluation}

\subsection{Datasets}
To fully demonstrate the effectiveness of our approach, we selected three popular visual localisation datasets that encompass indoor scenes, large-scale outdoor scenes, and challenges involving day-night changes, moving objects, and motion blur. The \textit{7Scenes} dataset~\cite{glocker2013real,shotton2013scene} is an indoor dataset consisting of seven small scenes from $1m^3$
to $18m^3$. The \textit{Cambridge Landmarks}~\cite{kendall2015posenet} dataset is a large-scale outdoor dataset with five scenes. The \textit{Aachen Day-Night-v1.1}~\cite{sattler2018benchmarking,zhang2021reference} dataset comprises 6697 training and 1015 test images taken from Aachen, Germany. The database images were captured during the daytime using handheld cameras, while query images were collected with three different mobile phones in both day and night conditions. The test set is split into two subsets: day and night. Nighttime images are only for testing.

\subsection{Implementation details}
\label{sec:hyper}
We integrate NetVLAD (NV) and EigenPlaces (EP) into \sysname and conduct visual localisation experiments because these two IR models have a stronger correlation with the match ratio than AP-GeM, as shown in Figure~\ref{fig:match_ratio_simi}. Different IR methods have an impact on similarity thresholds. As shown in Figure~\ref{fig:match_ratio_simi}, although the correlation between the three IR methods is similar for each dataset, the range distribution of similarity is slightly different. For EigenPlaces, the distribution of similarity on all datasets is 0.3 to 0.95, but NetVLAD and AP-GeM are not so uniform. For NetVLAD in particular, the range of similarity on the Aachen Day-Night-v1.1 dataset is smaller than on the Cambridge Landmarks and 7Scenes datasets. There are two possible reasons for this phenomenon. Firstly, the Aachen-V1.1 dataset represents a handheld motion pattern for AR applications in large city-scale environments, and EigenPlaces is more robust to camera viewpoint changes than NetVLAD and AP-GeM~\cite{Berton_2023_EigenPlaces}. Secondly, the Aachen Day-Night-v1.1 dataset has a cross-device domain gap, as the reference and query images have different resolutions and camera intrinsic parameters, whereas the images in the 7Scenes and Cambridge datasets were captured with the same camera.

Therefore, for NetVLAD, we set $\gamma^l = 0.4$ and $\gamma^h = 0.6$ for Cambridge and 7Scenes datasets, $\gamma^l = 0.3$ and $\gamma^h = 0.5$ for the Aachen-V1.1 dataset.  For EigenPlaces, we set $\gamma^l = 0.7$ and $\gamma^h = 0.9$ for all three datasets. We set $\alpha = 0.5$ and $\beta = 0.7$ for all IR methods and all three datasets.  We conduct experiments based on the HLoc toolbox~\cite{sarlin2019coarse}.%We utilize the structure from motion (SfM) model provided in HLoc official code\footnote{\url{https://github.com/cvg/Hierarchical-localisation/tree/master/hloc/pipelines/}} for three datasets.

\subsection{Evaluation}We evaluate the performance of HLoc and \sysname through two primary metrics. 
We consider the mean and median absolute translation error (ATE) and absolute rotation error (ARE) for all test frames in Cambridge Landmarks and 7Scenes datasets. For Aachen Day-Night-v1.1, we analyze the percentage of test images with pose predicted with high ($0.25m, 2^\circ$), medium ($0.5m, 5^\circ$), and low ($5m, 10^\circ$) accuracy levels proposed by~\cite{sattler2018benchmarking}\footnote{This dataset does not provide ground truth poses for each test frame.}.

\subsection{Results}
\begin{table}[ht]
%\footnotesize
\centering
\caption{Comparisons on Cambridge Landmarks dataset.  Median translation and rotation errors (cm/$^\circ$). We report the results of HLoc and \sysname using $k=10$.}
\setlength{\tabcolsep}{1pt} %
%\resizebox{0.5\textwidth}{!}{$\displaystyle
\begin{threeparttable}
\resizebox{1\columnwidth}{!}{
\begin{tabular}{l|c|cccc|c}
\toprule
 &Methods & Kings  & Hospital  & Shop & Church  &Avg. $\downarrow$ [$\text{cm}/^\circ$]   \\\midrule
 \multirow{4}{*}{APR}&PoseNet~\cite{kendall2015posenet} & 93/2.73 & 224/7.88 & 147/6.62 & 237/5.94 & 175/5.79 \\
&MS-Transformer~\cite{shavit2021learning}  & 85/1.45 &175/2.43& 88/3.20 & 166/4.12 & 129/2.80 \\
&LENS~\cite{moreau2022lens} & 33/0.5 &44/0.9& 27/1.6 & 53/1.6 & 39/1.15 \\
&DFNet~\cite{chen2022dfnet} & 73/2.37 & 200/2.98 & 67/2.21 & 137/4.02 & 119/2.90 \\
\midrule
\multirow{3}{*}{SCR}& ACE~\cite{brachmann2023accelerated} & 29/0.38 & 31/0.61 & 5/0.3 & 19/0.6 & 21/0.47 \\
%& ACE Ensemble~\cite{brachmann2023accelerated} & 18/0.3&25/0.5&5/0.3&9/0.3& 14/0.35 \\
& DSAC*~\cite{brachmann2021visual} & 18/0.3&21/0.4&5/0.3&15/0.6& 15/0.4 \\
&GLACE~\cite{wang2024glace} & 19/0.32 & 18/0.42 & 5/0.22 & 9/0.3 & 13/0.32\\
\midrule
\multirow{3}{*}{SOTA}& PixLoc~\cite{sarlin2021back} &14/\textbf{0.2}& 16/\textbf{0.3} &5/\textbf{0.2}& 10/0.3 & 11.3/0.25\\
&HLoc (SP+SG)~\cite{sarlin2020superglue,detone2018superpoint} &11.3/\textbf{0.2} &  15/0.31 & \textbf{4.2}/0.21 & \textbf{7.1}/0.22&9.4/0.24\\
& \textbf{\sysname (SP+SG) (ours)} & \textbf{11.2/0.2} & \textbf{14.6}/0.32 & \textbf{4.2/0.20} & 7.3/\textbf{0.20} &  
\textbf{9.3/0.23}\\
\bottomrule 
\end{tabular}
}
\end{threeparttable}
\label{tab:acc_cam}
\end{table}
We present the localisation accuracy of HLoc and \sysname across three distinct datasets, as depicted in Figure~\ref{fig:error_k} and Figure~\ref{fig:error_k_aachen}, with respect to varying values of $k$. In the case of HLoc, we report the results for $k = 1, 2, 3, 4, 5, 10, 20, 30$ and it retrieves the same number of $k$ images from the database for all queries. For \sysname, we provide the outcomes for $k = 4, 5, 10, 20, 30$, considering a retrieval of top-$k$ similar images solely for hard queries, top-$k^*$, $k^* = \lceil \alpha\times k \rceil$ images for easy queries, and top-$k^*$, $k^* = \lceil \beta\times k \rceil$ images for medium queries. We do not give the results for $k\leq3$ because the optimization space is too small and most datasets achieve much higher accuracy when $k\geq 10$.
\subsubsection{Cambridge Landmarks} 
For both HLoc and \sysname, the mean and median pose errors converge when $k \geq 10$. As shown in Figures~\ref{fig:error_k} (a)-(b) and Figure~\ref{fig:ir_ratio_runtime} (a), \sysname achieves nearly the same accuracy as HLoc while retrieving up to 30\% fewer images. The performance of both IR modules is comparable. Table~\ref{tab:acc_cam} highlights that \sysname achieves SOTA accuracy compared to other methods.

\subsubsection{7Scenes} 
\begin{figure}
\centering
\subfloat[Cam. ARE]{
 %% label for second subfigure
\includegraphics[width=.45\linewidth]{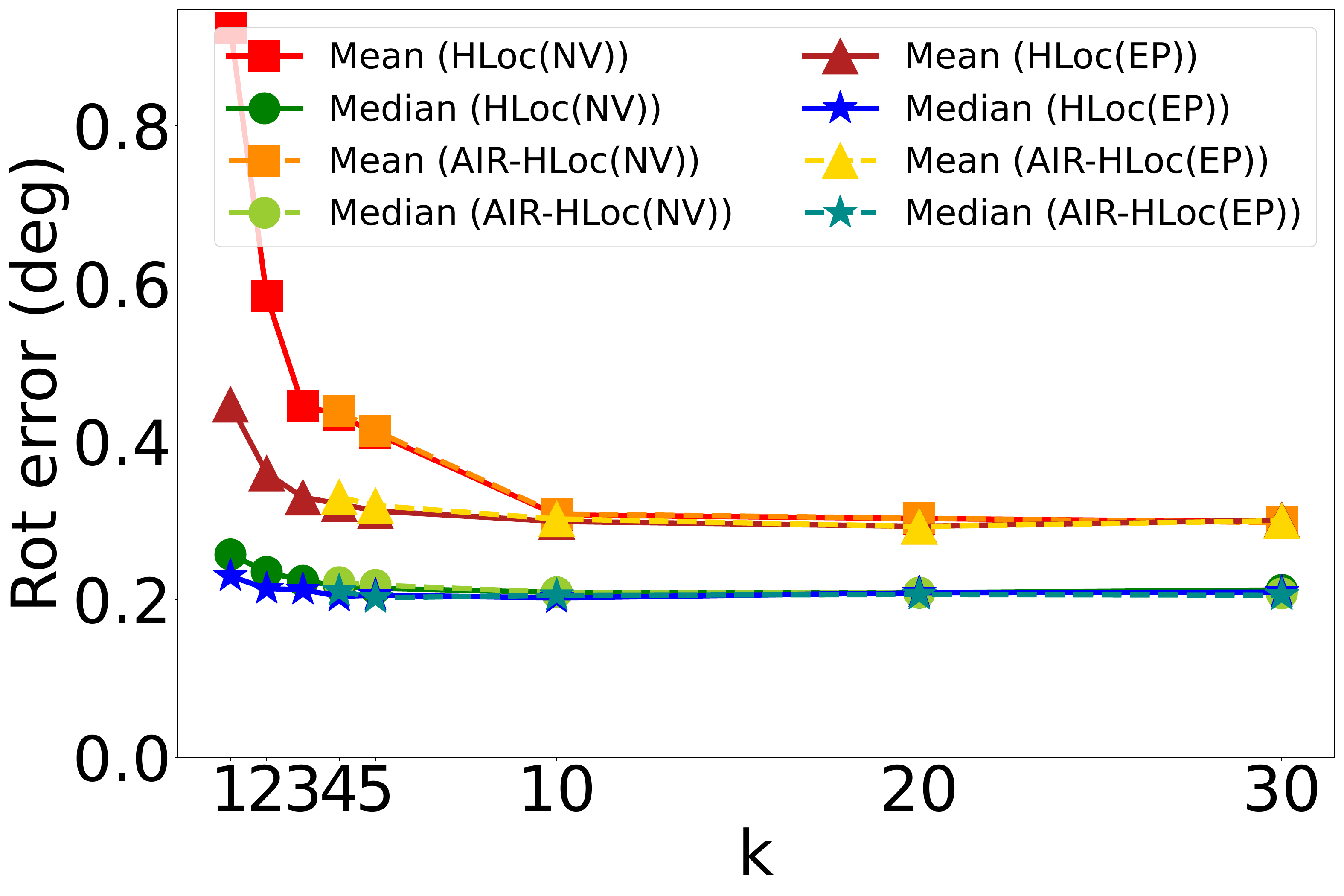}}
\subfloat[Cam. ATE]{
 %% label for second subfigure
\includegraphics[width=.45\linewidth]{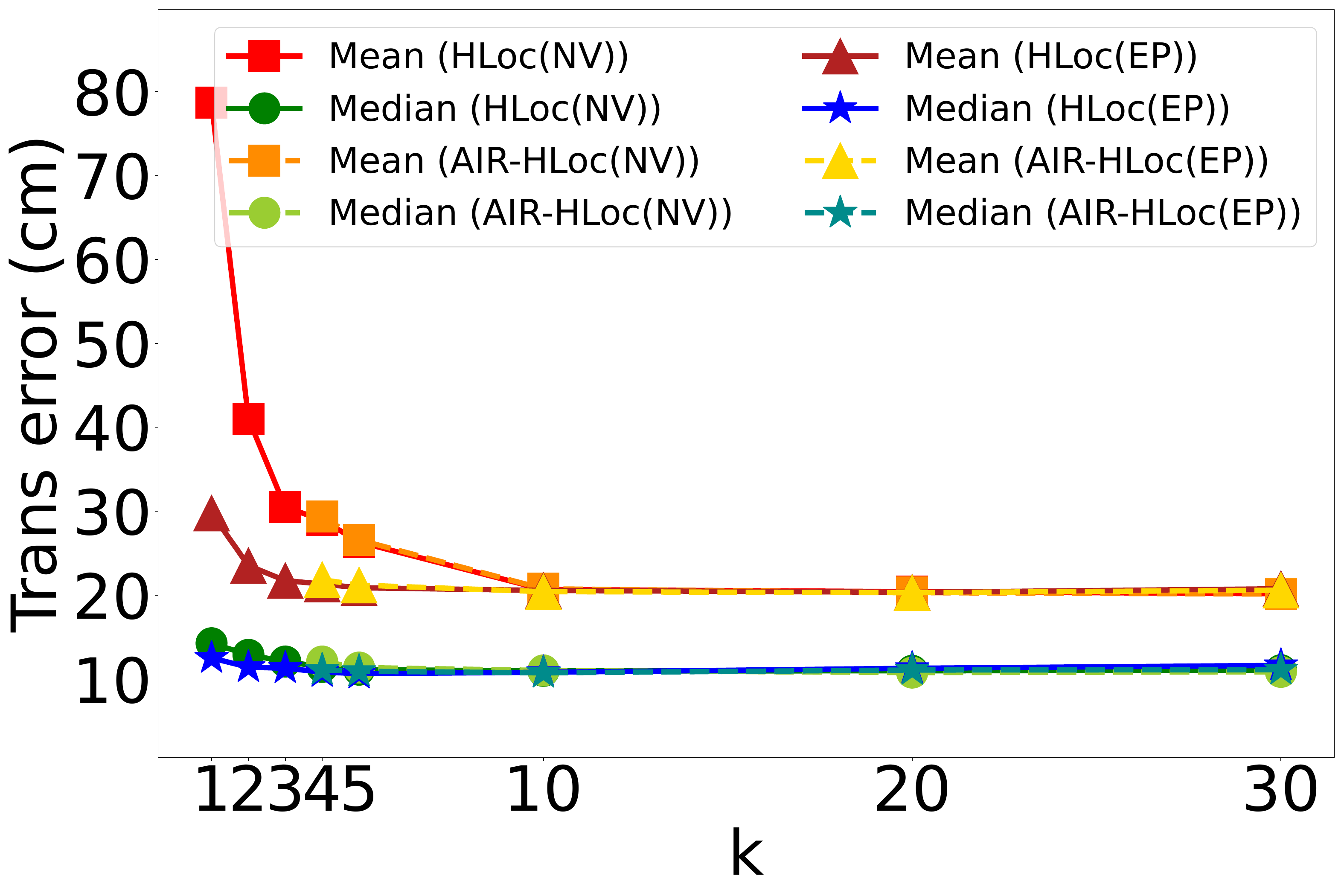}}
\newline
\subfloat[7S. ARE]{
 %% label for second subfigure
\includegraphics[width=.45\linewidth]{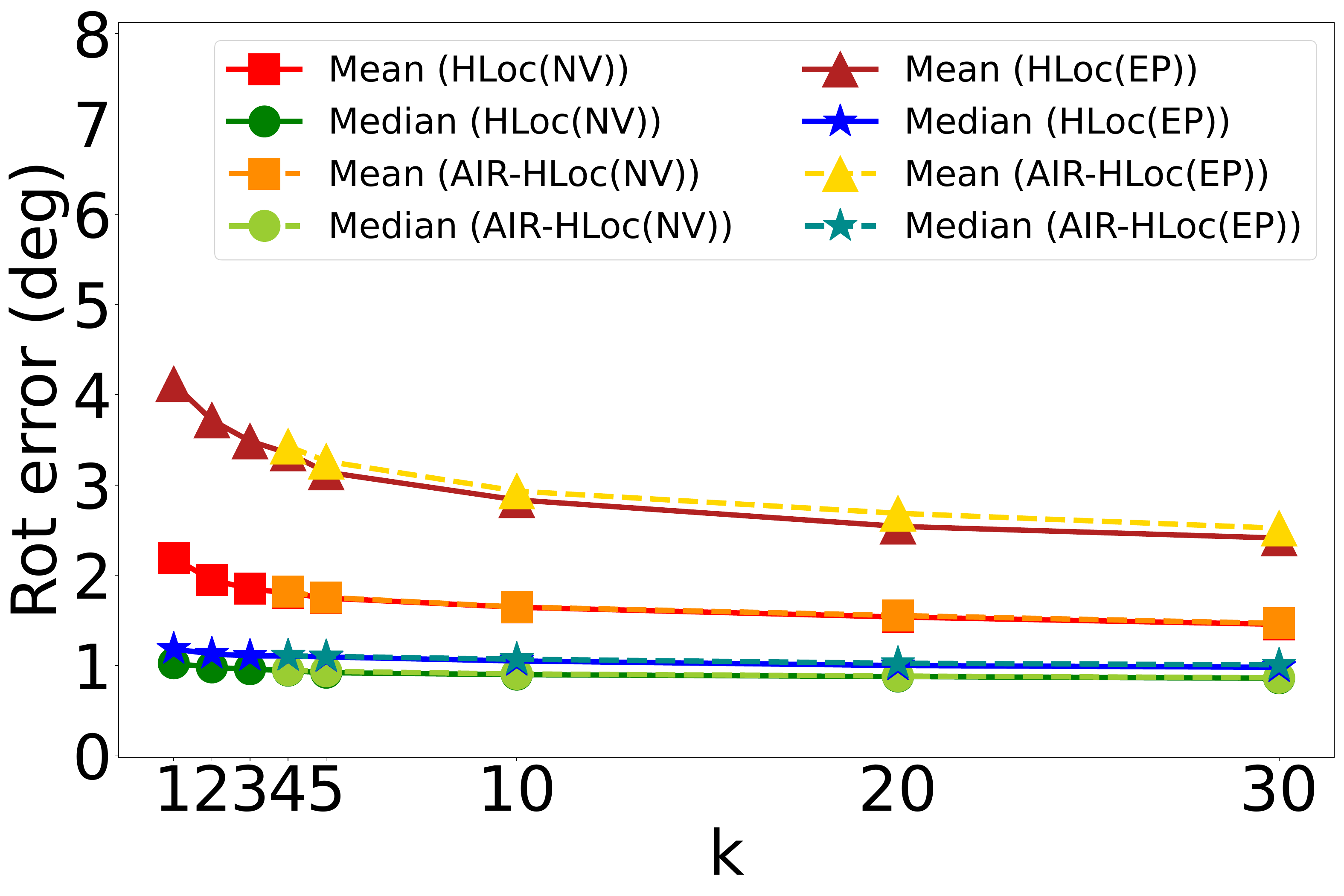}}
\subfloat[7S. ATE]{
 %% label for second subfigure
\includegraphics[width=.45\linewidth]{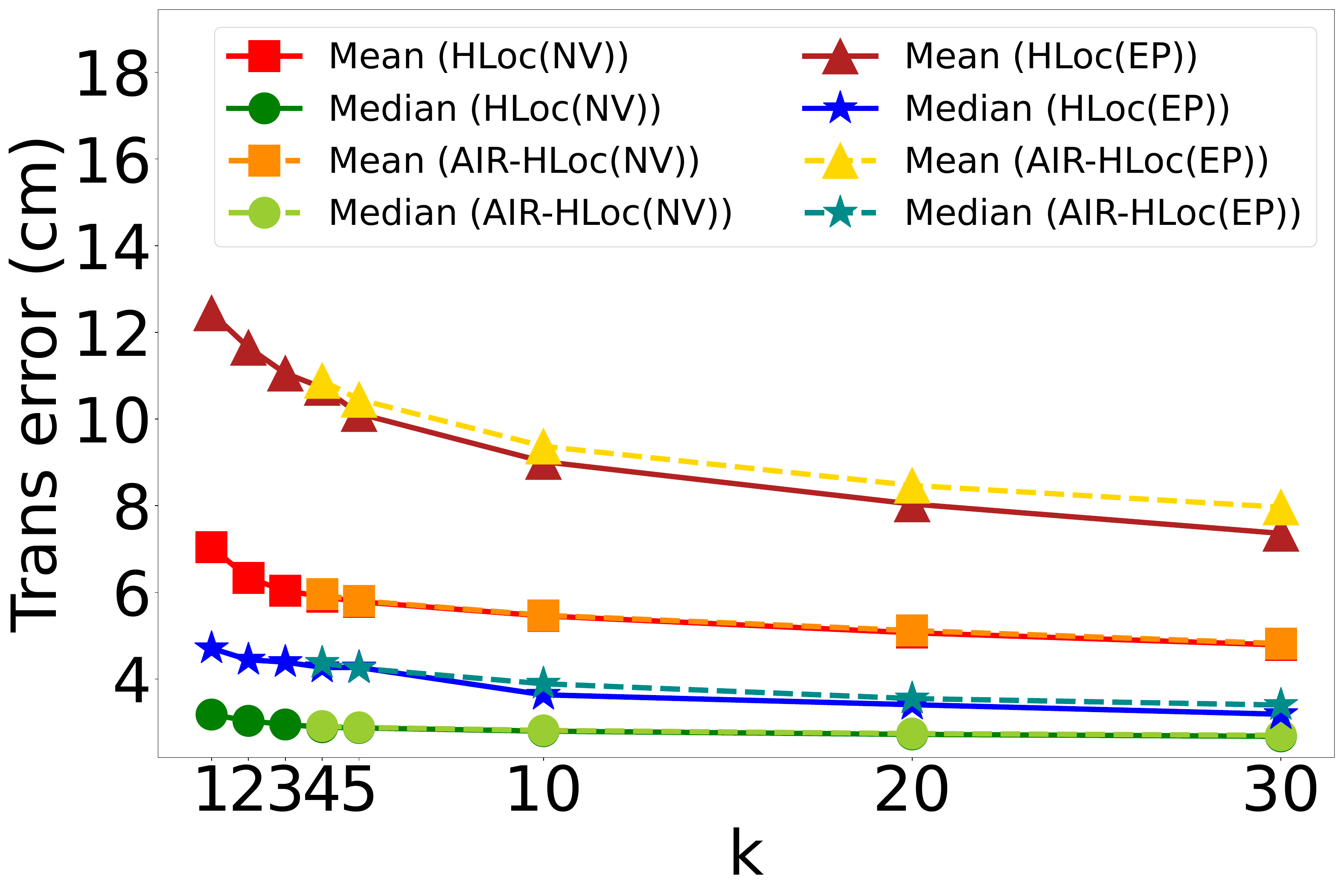}}
 \caption{Average mean and median pose error (ATE, ARE) for Cambridge and 7Scenes datasets across all scenes against $k$. For HLoc, it retrieves top-$k$ similar images for all queries. For \sysname, it retrieves $k$ similar images only for hard queries and $k^*$ images for medium and easy queries. AIR-HLoc (NV) uses NetVLAD as the image retrieval module, while AIR-HLoc (EP) utilizes EigenPlaces for image retrieval. The average \textit{retrieved ratio} ($k^*/k$) for all queries is shown in Figure~\ref{fig:ir_ratio_runtime} (a).}
 \label{fig:error_k} %% label for entire figure
\end{figure}

Similarly, for both HLoc and \sysname, the mean and median pose errors converge when $k \geq 20$. Figures~\ref{fig:error_k} (c)-(d) and Figure~\ref{fig:ir_ratio_runtime} (a) demonstrate that \sysname matches HLoc's accuracy while retrieving up to 26\% fewer images. Two IR modules exhibit similar performance. Table~\ref{tab:acc_7s} shows that \sysname and HLoc achieve SOTA accuracy compared to other approaches.

\begin{figure}[!ht]
\centering
% 插入图例，并去掉子图标记
\includegraphics[width=.75\linewidth]{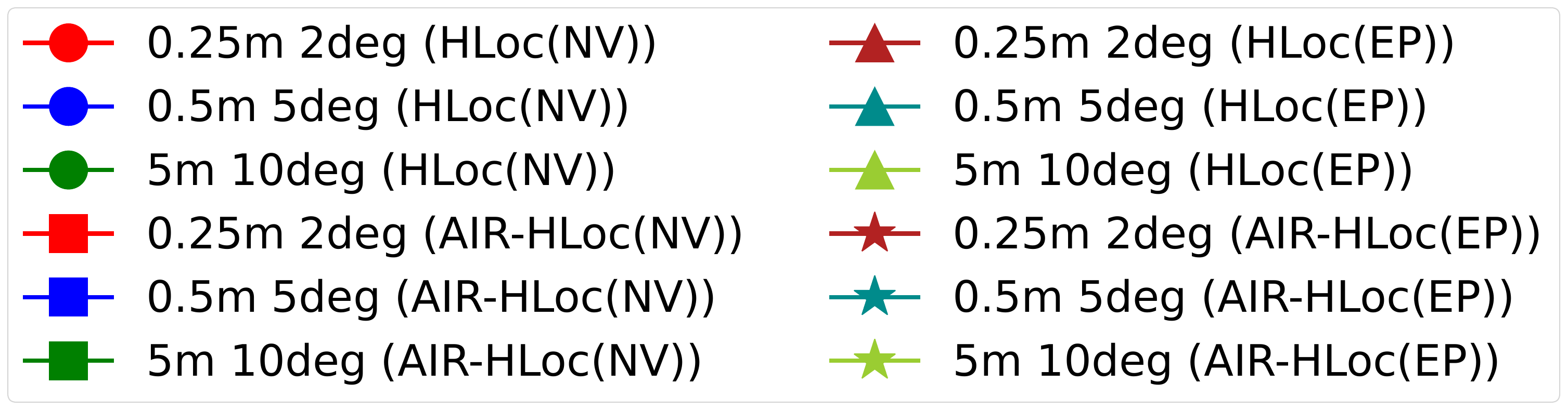} 
\vspace{-0.5em} % 调整图例与下面子图的垂直间距
% 插入两个子图
\subfloat[AachenV1.1, Day]{
\label{fig:aachen_day} %% 唯一的label
\includegraphics[width=.43\linewidth]{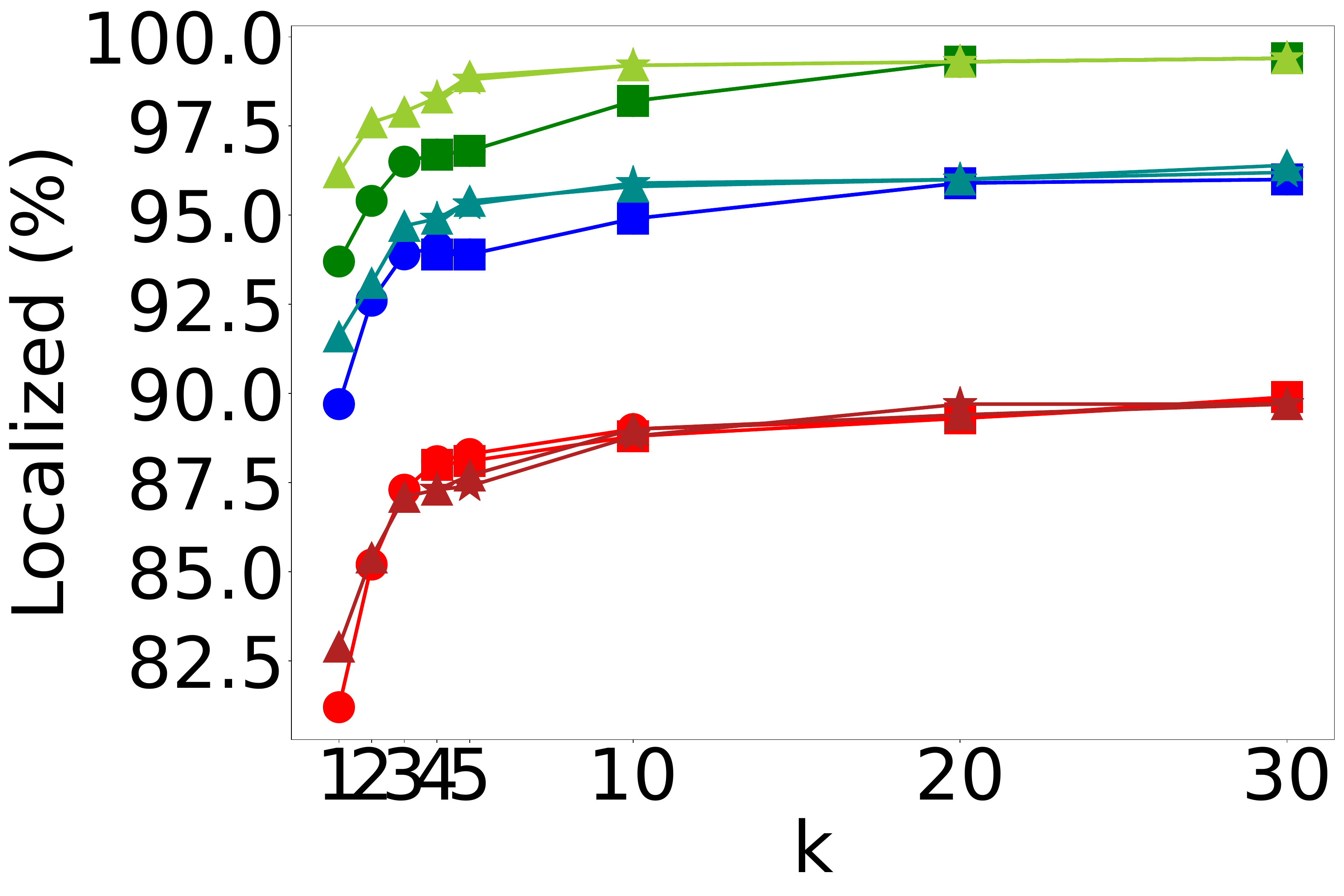}}
\subfloat[AachenV1.1, Night]{
\label{fig:aachen_night} %% 唯一的label
\includegraphics[width=.43\linewidth]{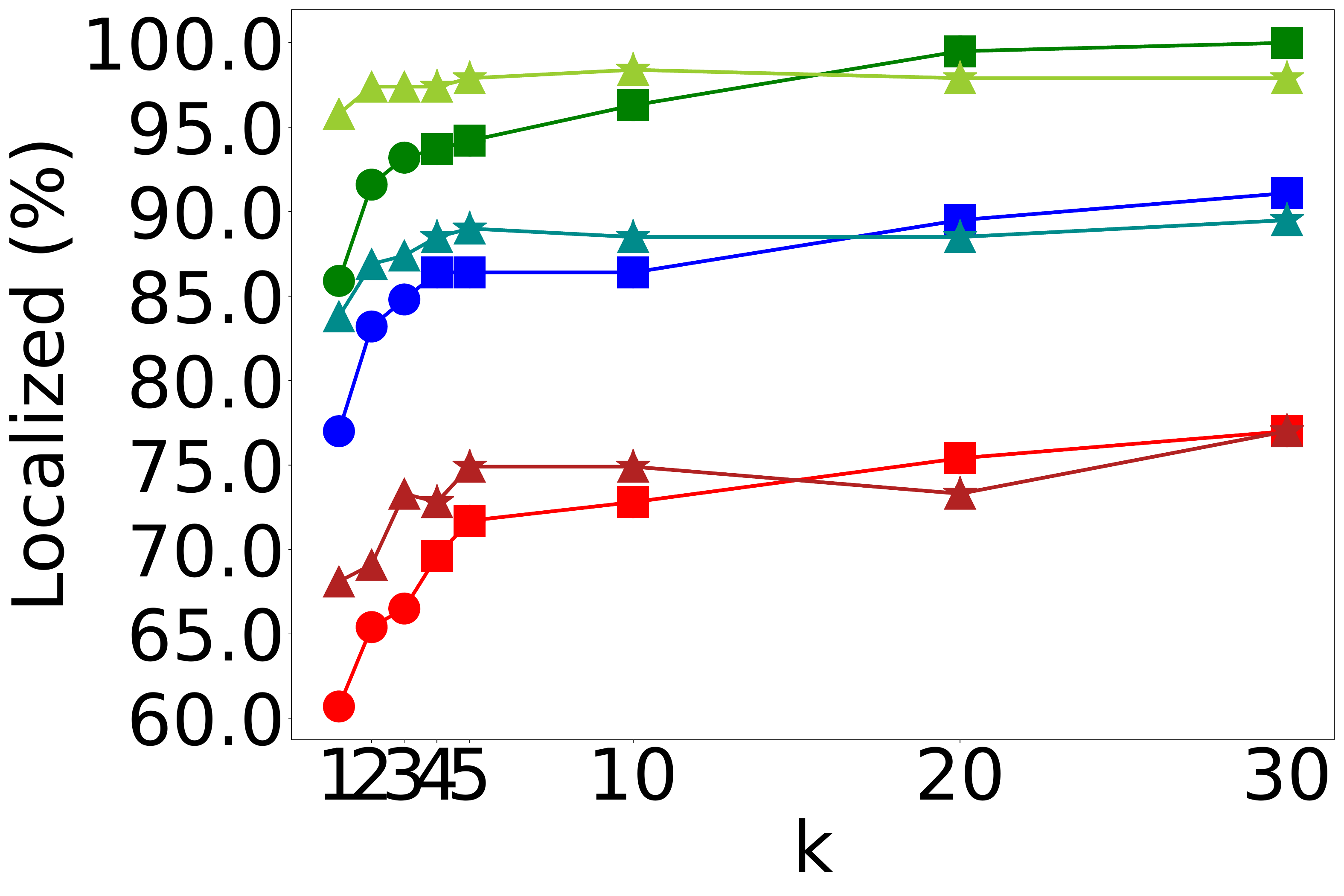}}
% 添加图标题
\caption{Percentage (\%) of test frames high (0.25m, $2^{\circ}$), medium (0.5m, $5^{\circ}$), and low (5m, $10^{\circ}$) accuracy~\cite{sattler2018benchmarking} (higher is better) for HLoc and \sysname against $k$. The average \textit{retrieved ratio} ($k^*/k$) is shown in Figure~\ref{fig:ir_ratio_runtime} (a).}
\label{fig:error_k_aachen} %% 总图label
\end{figure}

\subsubsection{Aachen Day-Night-v1.1} 
\begin{figure}[h!]
\centering
\subfloat[Retrieved Ratio]{
\includegraphics[width=0.48\linewidth, height=0.25\textheight, keepaspectratio]{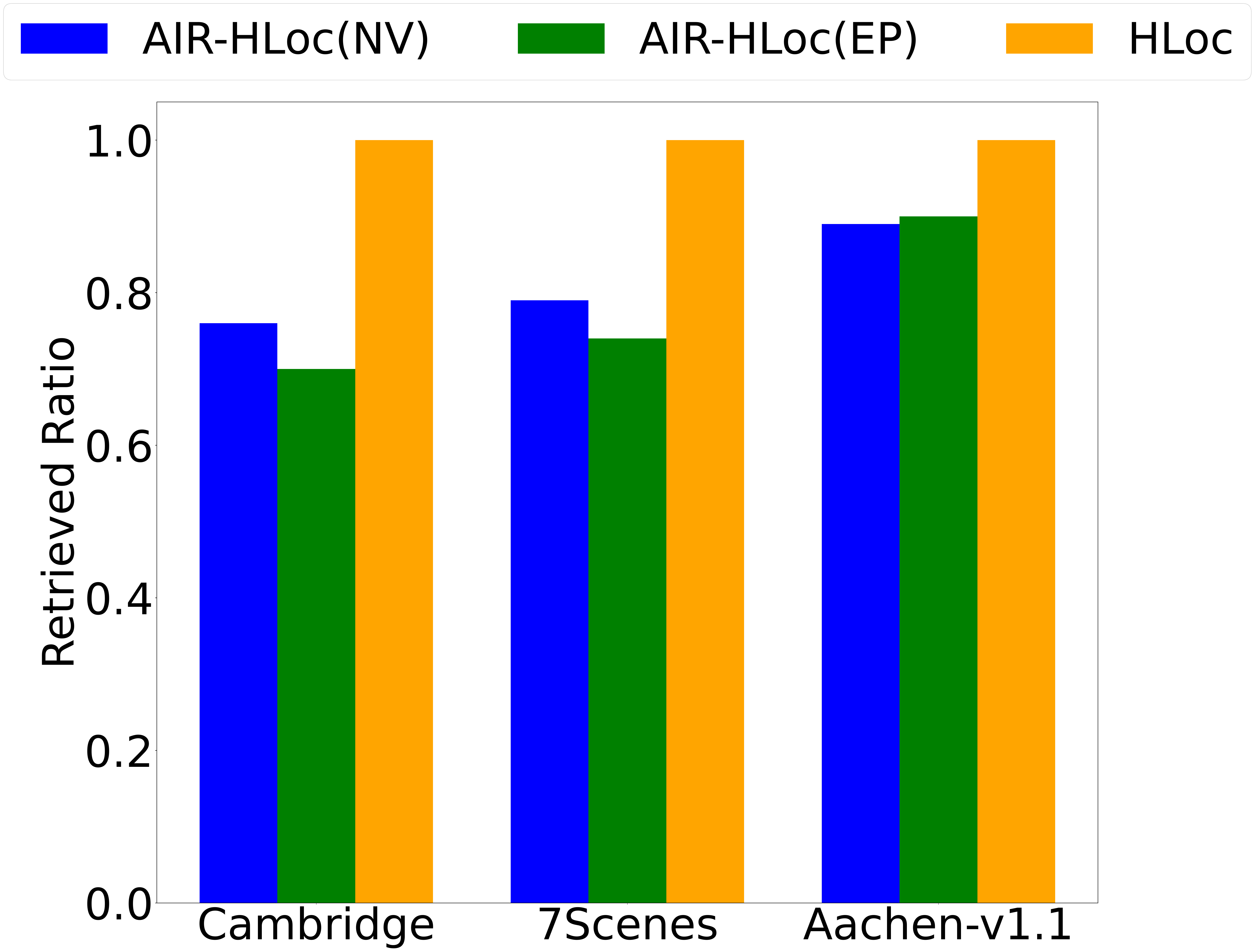}}
\subfloat[Feature Matching Time (SP+SG)]{
\includegraphics[width=0.48\linewidth, height=0.4\textheight, keepaspectratio]{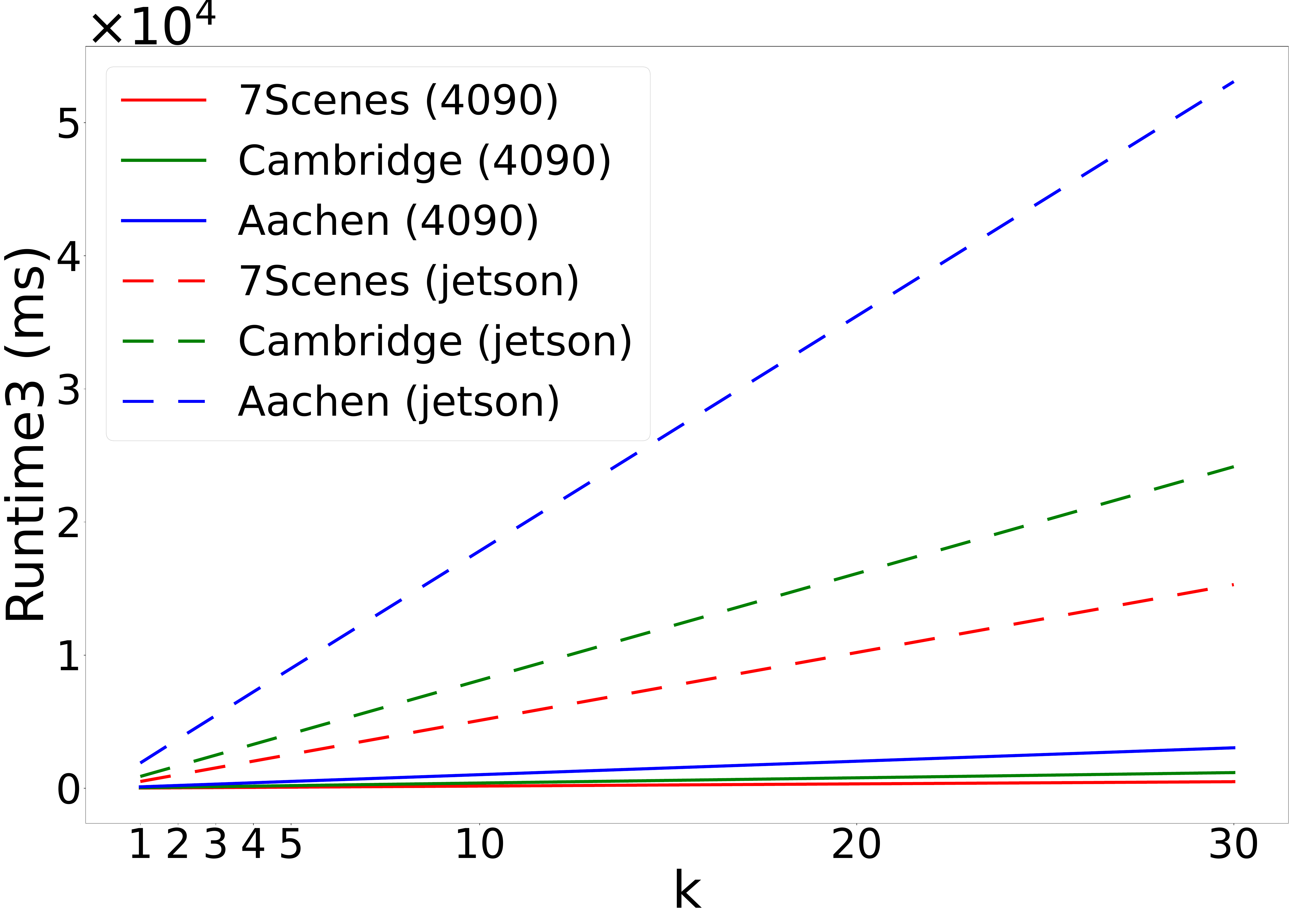}}
% 添加图标题
\caption{(a) \textit{Retrieved Ratio} refers to the ratio ($0<k^*/k\leq1$) of the average number of retrieved images for AIR-HLoc compared to HLoc for all test frames. (b) The feature matching time (runtime3) of HLoc in SuperPoint (SP) + SuperGlue (SG) setting for three datasets against $k$.}  %We keep the maximum number of feature points for each query $\leq$ 4096. The average runtime3 for each pair are 41ms, 120ms, and 227ms, respectively, for 7Scenes, Cambridge and Aachen-V1.1 when $k=10$.}
\label{fig:ir_ratio_runtime} %% 总图label
\end{figure}

For both HLoc and \sysname, the three accuracy levels converge when $k\geq 20$. Subfigures~\ref{fig:error_k_aachen} (a)-(b) and Figure~\ref{fig:ir_ratio_runtime} (a) show that \sysname achieves almost the same accuracy as HLoc with up to 11\% fewer retrieved images. We find that \sysname achieves accuracy comparable to that of HLoc. \sysname (EP) achieves slightly better accuracy than HLoc (EP) in the (0.25m, $2^{\circ}$) range when $k = 30$ in daytime test sequences. This demonstrates that \sysname can improve accuracy by reducing the noise introduced by a high $k$ value.

\subsection{Analysis}
\begin{figure*}
\centering
\subfloat[ATE, NetVLAD]{
 %% label for second subfigure
\includegraphics[height=1.1in, width=.24\linewidth]{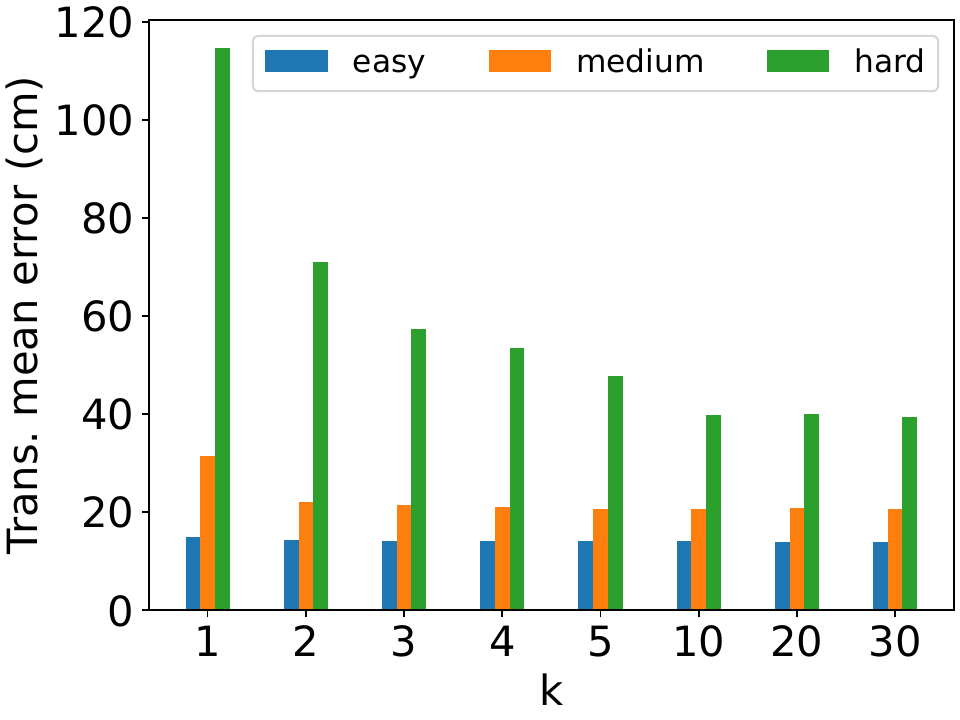}}
\subfloat[ARE, NetVLAD]{
 %% label for second subfigure
\includegraphics[height=1.1in, width=.24\linewidth]{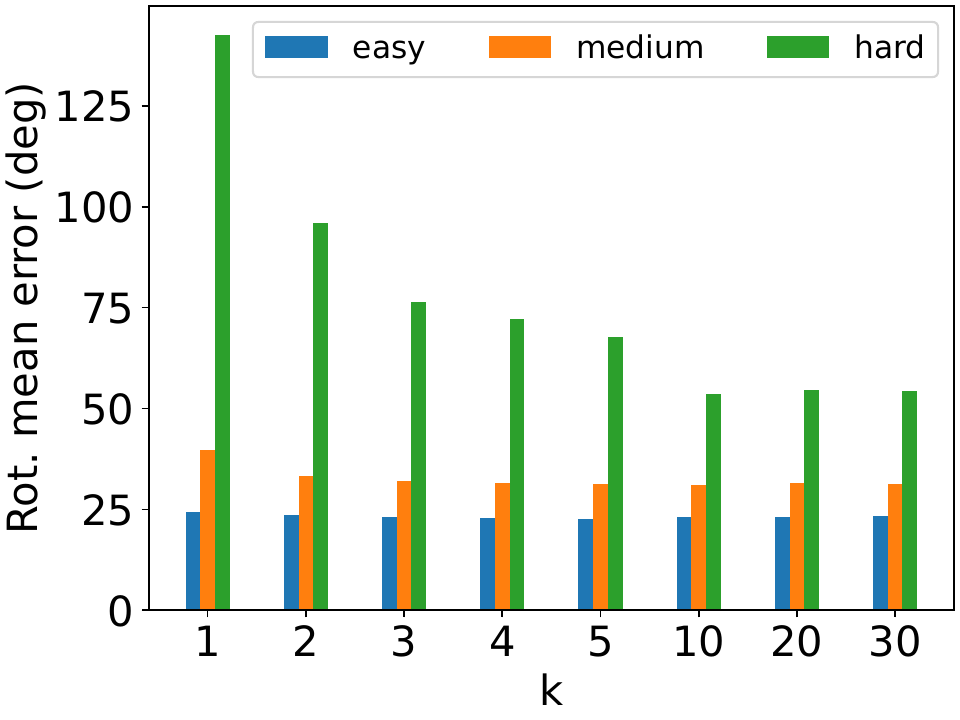}}
\subfloat[$\zeta_T(k)$, NetVLAD]{
 %% label for second subfigure
\includegraphics[height=1.1in, width=.24\linewidth]{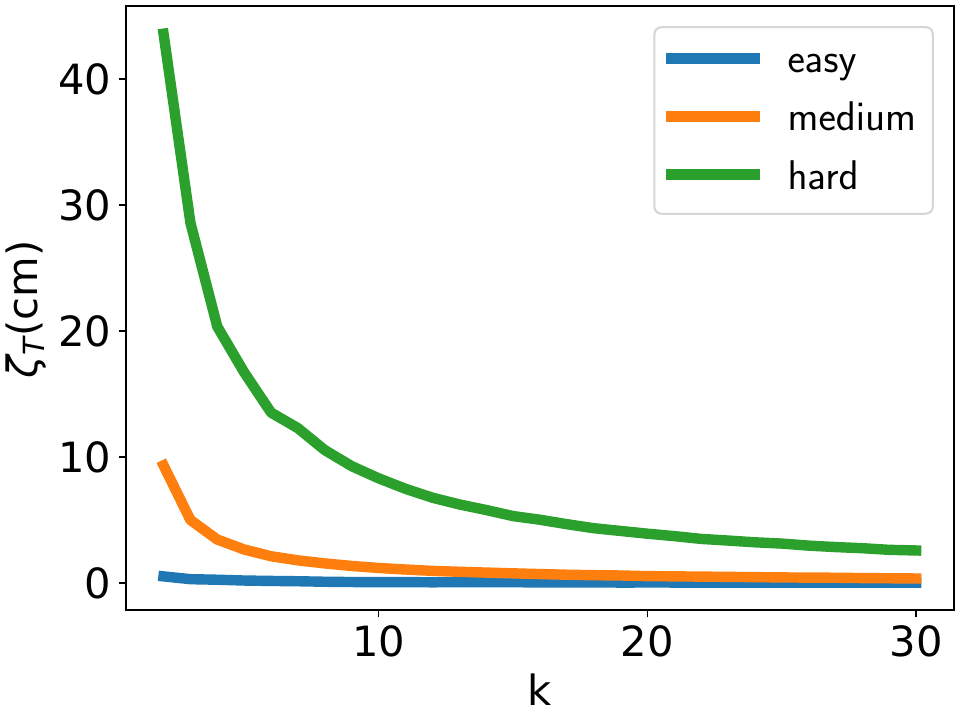}}
\subfloat[$\zeta_R(k)$, NetVLAD]{
 %% label for second subfigure
\includegraphics[height=1.1in, width=.24\linewidth]{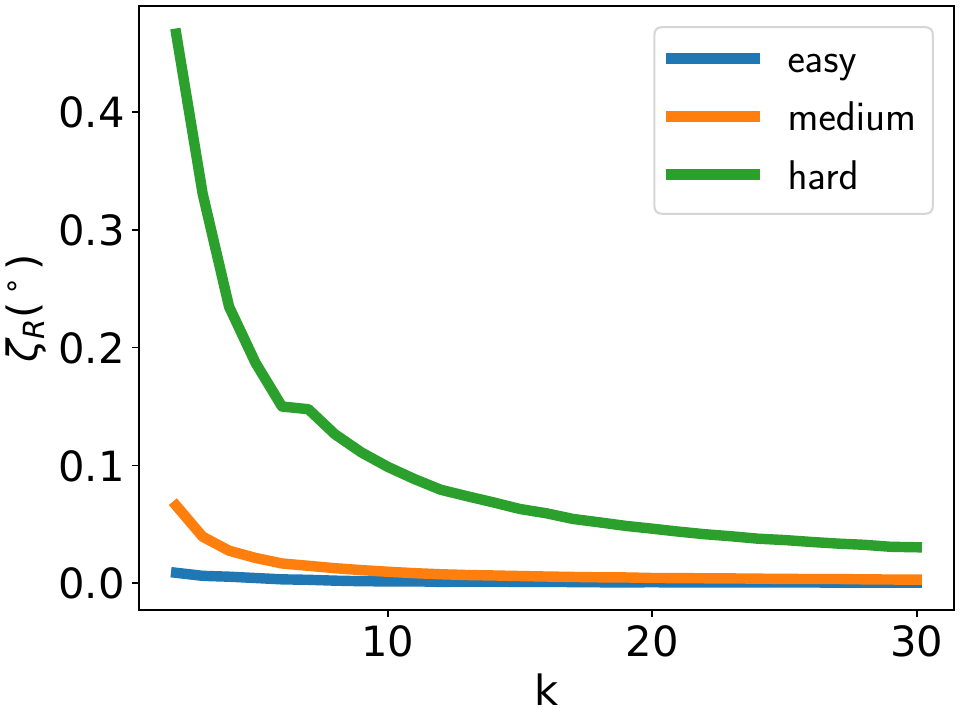}}
 \caption{ (a)- (b) Plots of mean rotation and translation errors of HLoc for easy, medium and hard queries of described in Section~\ref{sec:method} against $k$ in Cambridge Landmarks. (c)- (d) The  average MLIP, \( \zeta = (\zeta_T(k), \zeta_R(k)) \) of all query frames against $k$ from 1 to 30 in Cambridge Landmarks.}
 \label{fig:simi_error_bar_rayr} %% label for entire figure
\end{figure*}

\begin{figure}[t]
  \centering
  \includegraphics[width=.95\linewidth]{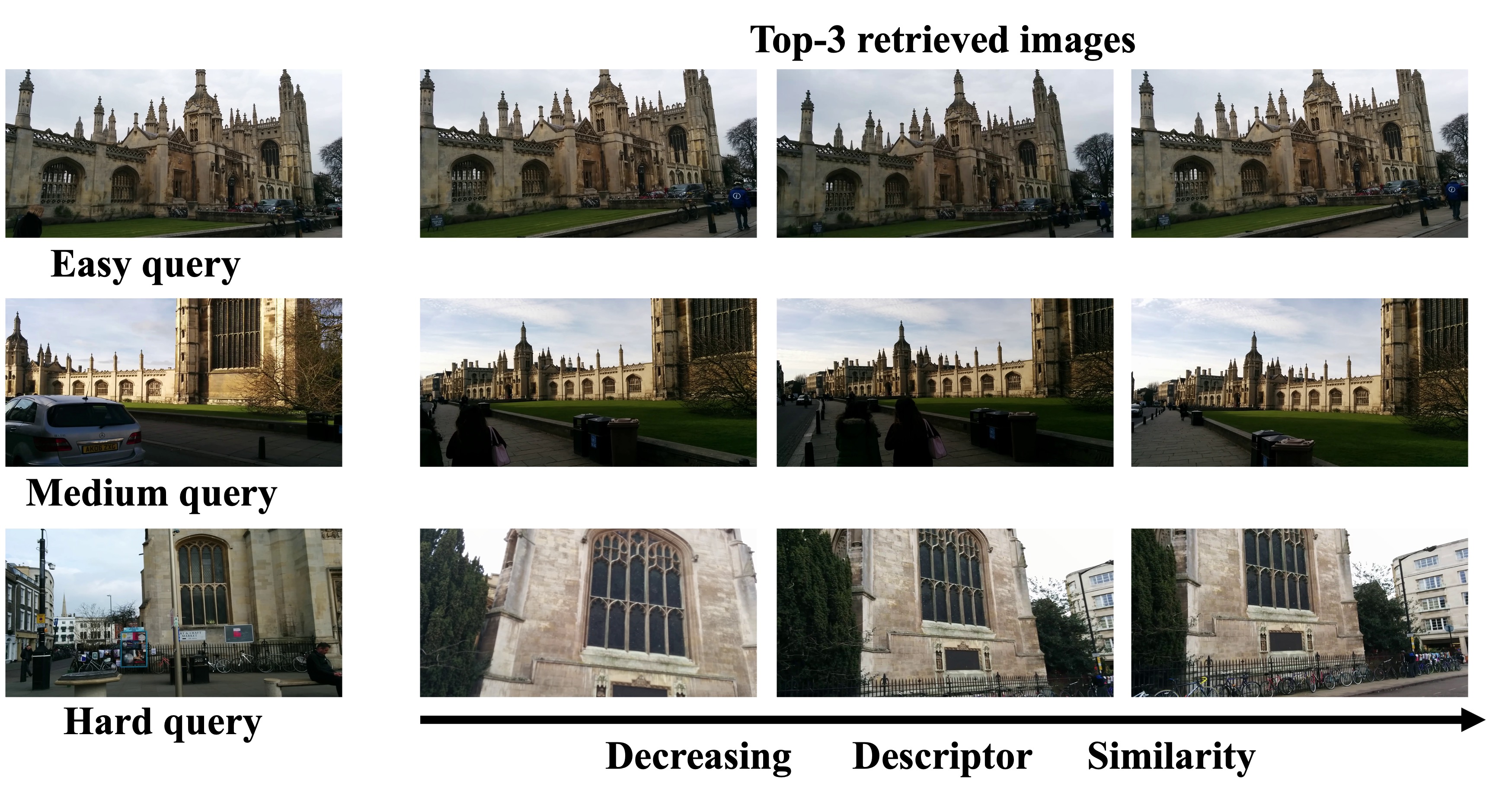}
  \caption{Examples of easy, hard, queries and corresponding top-3 retrieved images in Cambridge Landmarks. Left side are query images in the test set, right side are reference images retrieved by the EigenPlaces model.}
\label{fig:query_example}
\end{figure}

We present examples of queries classified as easy, medium, and hard alongside their top three retrieved images in  Figure~\ref{fig:query_example}. Easy query images are very similar to retrieved images, and there are relatively similar images for medium queries. In contrast, for hard queries, the retrieved images are even incorrect, showing no overlap with the query images. As depicted in Figure~\ref{fig:simi_error_bar_rayr} (a)-(d), increasing $k$, primarily improves the accuracy for hard queries. For easy and medium queries, further gains become negligible when $k \geq 5$. Furthermore, the average contribution of the top-$k$ images to the overall localisation accuracy converges to zero when $k \geq 10$ for these query types. As $k$ increases, the reduction in pose estimation error is primarily observed for hard queries, while easy and medium queries show very limited improvement.

Figures~\ref{fig:simi_error_bar_rayr} (a)-(b) demonstrate that pose estimation for easy queries is significantly more accurate compared to hard queries, even when $k$ is sufficiently large. This observation suggests that the similarity score of a query is indicative of the uncertainty in pose estimation. These results highlight the inefficiency of using a fixed number of retrieved images for feature matching across all queries, underscoring the effectiveness of \sysname.

\begin{table*}[ht]
\caption{Comparisons on 7Scenes dataset.  The median translation and rotation errors (cm/$^\circ$) of different methods. The best results are in bold (lower is better). We report the results of HLoc and \sysname using $k=30$.}
\centering
\setlength{\tabcolsep}{4pt}
\begin{threeparttable}
\resizebox{2\columnwidth}{!}{
\begin{tabular}{l|c|ccccccc|c}
\toprule    & Methods& Chess  & Fire  & Heads & Office &Pumpkin  & Redkitchen& Stairs  & Avg. $\downarrow$ [$\text{cm}/^\circ$] \\
\midrule
 \multirow{4}{*}{APR}& PoseNet~\cite{kendall2015posenet} & 32/8.12 & 47/14.4& 29/12.0 & 48/7.68& 47/8.42& 59/8.64 &  47/13.8 &  44/10.4\\
 & MS-Transformer~\cite{shavit2021learning}& 11/4.66 &24/9.60 &14/12.2 &17/5.66& 18/4.44&  17/5.94&  17/5.94 & 18/7.28 \\
 & DFNet~\cite{chen2022dfnet}& 5/1.88 &17/6.45 &  6/3.63& 8/2.48
&10/2.78 & 22/5.45 &  16/3.29 & 12/3.71  \\
 & Marepo~\cite{chen2024map}& 2.6/1.35 &2.5/1.42 &2.3/2.21 & 3.6/1.44&  4.2/1.55
& 5.1/1.99 & 6.7/1.83 & 3.9/1.68
 \\\midrule
\multirow{2}{*}{SCR} & DSAC*~\cite{brachmann2021visual}& \textbf{1.9}/1.1 &\textbf{1.9}/1.2 &  1.1/1.8& \textbf{2.6}/1.2&4.2/1.4 &  \textbf{3.0}/1.7& 4.2/1.4 & \textbf{2.7}/1.4 \\
 & ACE~\cite{brachmann2023accelerated}& \textbf{1.9/0.7} &\textbf{1.9}/0.9 & \textbf{0.9/0.6}&  2.7/\textbf{0.8}& 4.2/1.1 & 4.2/1.3 & \textbf{3.9}/1.1 & 2.8/0.93 \\\midrule
\multirow{2}{*}{SOTA}& HLoc (SP+SG) & 2.1/\textbf{0.7} & \textbf{1.9/0.8}&1.1/0.7 & \textbf{2.6/0.8} & \textbf{3.9/1.0} & 3.2/\textbf{1.1} & \textbf{3.9/1.0} & \textbf{2.7/0.87} \\
& \textbf{\sysname (SP+SG)(ours)} & 2.2/\textbf{0.7} & \textbf{1.9/0.8}& 1.1/0.7& \textbf{2.6/0.8} & \textbf{3.9/1.0} & 3.2/\textbf{1.1} &  \textbf{3.9/1.0} & \textbf{2.7/0.87} \\
\bottomrule
\end{tabular}
}
\end{threeparttable}
\label{tab:acc_7s}
\end{table*}
\subsection{System Efficiency and Application}
\label{subsec:se}
We first conduct the runtime analysis using an NVIDIA GeForce RTX 4090 GPU on a desktop. To illustrate the bottleneck introduced by feature matching and its impact on mobile robot localisation tasks, we also evaluate the runtime on an embedded NVIDIA Jetson Orin platform with \(64\) GB RAM. \ul{4090}: We assess the runtime of the entire HLoc pipeline. Since global and local feature extraction for the reference image database can be performed offline, we focus on the runtime for query image feature extraction (runtime1), image retrieval (runtime2), feature matching (runtime3), and pose estimation (runtime4). Runtime1 and runtime4 are nearly constant per query, while runtime2 depends on the size of the image database. For King's College, runtime1 + runtime2 + runtime4 is approximately 40 ms per query. We observe that runtime3 increases linearly with \( k \), as shown in Figure~\ref{fig:ir_ratio_runtime} (b). Runtime3 is longer for Cambridge and Aachen-V1.1 compared to 7Scenes due to the higher resolution of images in these datasets\footnote{We resize Aachen Day-Night-v1.1 images to a maximum size of 1600 pixels and Cambridge landmarks images to 1024 pixels. For 7Scenes, we retain the original resolution of 640x480 pixels.}. The average runtime3 per image pair is 17 ms for 7Scenes, 40 ms for Cambridge, and 107 ms for Aachen-V1.1. Runtime3 dominates the pipeline's execution time when \( k \geq 10 \), while most datasets require at least \( k = 10 \) for optimal performance. At \( k = 10 \), \sysname reduces the average runtime3 across all queries by 44 ms, 120 ms, and 110 ms for 7Scenes, Cambridge, and Aachen-V1.1, respectively. \ul{Jetson Orin:} The average runtime3 per image pair is 510 ms for 7Scenes, 811 ms for Cambridge, and 1900 ms for Aachen-V1.1. At \( k = 10 \), \sysname reduces the average runtime3 across all queries by 1326 ms, 2433 ms, and 2090 ms for 7Scenes, Cambridge, and Aachen-V1.1, respectively. It shows that our \sysname significantly reduces the computational time on a mobile robot platform.

Modern robots often utilize visual-inertial odometry (VIO) to maintain local camera pose tracking~\cite{yu2022improving, bao2022robust, liu2024mobilearloc}. Absolute pose estimation is primarily needed for initial alignment and periodic recalibration of the tracking system. Given that VIO systems are increasingly resilient to drift, they can skip isolated difficult queries, relying instead on easier ones with higher accuracy and lower runtime. This approach, combined with \sysname, further reduces the computational load for real-time camera relocalisation while maintaining tracking accuracy.

\section{CONCLUSIONS}
 The IR module in HLoc retrieves the top-$k$ similar reference images for a given query. A larger $k$ improves localisation robustness for challenging queries but increases the computational cost, as feature matching— the main runtime bottleneck—scales linearly with $k$. This paper introduces \sysname, a method that optimizes processing time by adaptively retrieving a different number of images based on the similarity between query images and the reference image database. Extensive experiments conducted on three datasets demonstrate the efficacy of our algorithm, achieving up to a 30\% reduction in matching cost while maintaining state-of-the-art (SOTA) accuracy. This paper provides novel insights into the relationship between global and local descriptors, offering a deeper understanding that can inform more effective $k$ selection and algorithm design for practical applications.


%%%%%%%%%%%%%%%%%%%%%%%%%%%%%%%%%%%%%%%%%%%%%%%%%%%%%%%%%%%%%%%%%%%%%%%%%%%%%%%%

%%%%%%%%%%%%%%%%%%%%%%%%%%%%%%%%%%%%%%%%%%%%%%%%%%%%%%%%%%%%%%%%%%%%%%%%%%%%%%%%

%%%%%%%%%%%%%%%%%%%%%%%%%%%%%%%%%%%%%%%%%%%%%%%%%%%%%%%%%%%%%%%%%%%%%%%%%%%%%%%%
%\section*{APPENDIX}


%\section*{ACKNOWLEDGMENT}

%The preferred spelling of the word ÒacknowledgmentÓ in America is without an ÒeÓ after the ÒgÓ. Avoid the stilted expression, ÒOne of us (R. B. G.) thanks . . .Ó  Instead, try ÒR. B. G. thanksÓ. Put sponsor acknowledgments in the unnumbered footnote on the first page.

%%%%%%%%%%%%%%%%%%%%%%%%%%%%%%%%%%%%%%%%%%%%%%%%%%%%%%%%%%%%%%%%%%%%%%%%%%%%%%%%

% \bibliographystyle{mybstfile} 
% \bibliography{IEEEabrv,mybibfile}
\bibliographystyle{IEEEtran} 
\bibliography{IEEEabrv,IEEEexample}

\end{document}